\documentclass{article}

\usepackage{PRIMEarxiv}

\usepackage[utf8]{inputenc} 
\usepackage[T1]{fontenc}    
\usepackage{hyperref}       
\usepackage{url}            
\usepackage{booktabs}       
\usepackage{amsfonts}       
\usepackage{nicefrac}       
\usepackage{microtype}      
\usepackage{lipsum}
\usepackage{fancyhdr}       
\usepackage{graphicx}       
\graphicspath{{media/}}     

\usepackage{adjustbox}
\usepackage{tabularx}
\usepackage{colortbl}
\usepackage{svg}
\usepackage{float}
\usepackage{multirow}
\usepackage{subcaption}
\usepackage{booktabs}
\usepackage{colortbl}
\usepackage[table]{xcolor}
\usepackage{longtable}
\usepackage{array}
\usepackage{siunitx}  
\usepackage{caption}
\usepackage{subcaption}
\usepackage{enumitem}
\usepackage{graphicx}
\usepackage{grffile}
\usepackage{amsmath}
\usepackage{booktabs}
\usepackage{multirow}
\usepackage{array}
\usepackage{xcolor}
\usepackage{graphicx}
\definecolor{darkgreen}{RGB}{0,100,0}

\usepackage[utf8]{inputenc}
\usepackage[T1]{fontenc}
\usepackage{geometry}
\usepackage{amsmath}
\usepackage{tcolorbox}
\tcbuselibrary{listings, breakable, skins, theorems}
\usepackage{listings}

\usepackage[utf8]{inputenc}
\usepackage[T1]{fontenc}
\usepackage{amsmath}
\usepackage{tcolorbox}
\usepackage{listings}

\usepackage{lmodern}
\usepackage{hyperref}

\newtcolorbox{promptbox}[1][]{%
  breakable,
  enhanced,
  colback=gray!5,
  colframe=black!80,
  boxrule=0.5pt,
  sharp corners,
  title=#1,
  fonttitle=\bfseries,
  listing only,
  listing options={
    basicstyle=\ttfamily\small,
    breaklines=true,
    breakatwhitespace=true,
    columns=flexible,
    keepspaces=true,
    showstringspaces=false,
    tabsize=2,
    literate={
      {~}{{\textasciitilde}}1
      {_}{{\_}}1
      {\{}{{\char`\{}}1
      {\}}{{\char`\}}}1
      {<}{{\textless}}1
      {>}{{\textgreater}}1
    }
  },
  width=\textwidth,
  left=2pt,
  right=2pt,
  top=4pt,
  bottom=4pt,
  before skip=10pt,
  after skip=10pt,
}

\usepackage{listings}

\lstdefinelanguage{json}{
  basicstyle=\ttfamily\small,
  numbers=none,
  showstringspaces=false,
  breaklines=true,
  frame=single,
  backgroundcolor=\color{gray!10},
  literate=
   *{0}{{{\color{blue}0}}}{1}
    {1}{{{\color{blue}1}}}{1}
    {2}{{{\color{blue}2}}}{1}
    {3}{{{\color{blue}3}}}{1}
    {4}{{{\color{blue}4}}}{1}
    {5}{{{\color{blue}5}}}{1}
    {6}{{{\color{blue}6}}}{1}
    {7}{{{\color{blue}7}}}{1}
    {8}{{{\color{blue}8}}}{1}
    {9}{{{\color{blue}9}}}{1}
    {:}{{{\color{black}:}}}{1}
    {,}{{{\color{black},}}}{1}
    {"}{{{\color{red}"}}}{1}
}
\usepackage{booktabs}
\usepackage{makecell}
\usepackage{tabularx}

\usepackage{booktabs}
\usepackage{colortbl}
\usepackage[table,xcdraw]{xcolor}
\usepackage{tabularx}
\usepackage{makecell}

\definecolor{lightgreen}{RGB}{210, 240, 210}

\title{LLM-FS-Agent: A Deliberative Role-based Large Language Model Architecture for Transparent Feature Selection}

\makeatletter
\renewcommand\@fnsymbol[1]{\ifcase#1\or 1\or 2\or 3\or 4\or 5\or 6\or 7\or 8\else\@ctrerr\fi}
\makeatother

\author{
  Bal-Ghaoui Mohamed \footnotemark[1]{} \\
  R\&D Department, 
  Audensiel Conseil \\ 
  Paris, France \\
  \texttt{m.bal-ghaoui@audensiel.fr} \\
  \And
  Sabri Fayssal \footnotemark[2]{}\\
  Ecole Centrale de Lyon\\
  Lyon, France\\
  \texttt{fayssalsabri4@gmail.com} \\
}

\begin{document}
\maketitle

\vspace{-1cm}
\begin{center}
    7 October 2025\\
    $^1$ Corresponding author: research idea, methodology, review, validation. \\
    $^2$ Contributing author: software development, AI implementation, writing. \\
\end{center}
\vspace{0.5cm}

\begin{abstract}
The pervasive challenge of high-dimensional data in Machine Learning pipelines often
compromises model interpretability and efficiency.
While Large Language Models (LLMs) have shown potential in Dimensionality Reduction (RD) through
Feature Selection (FS), existing LLM-based approaches often lack structured reasoning and
transparent justification for their decisions. This paper introduces LLM-FS-Agent, a
novel multi-agent architecture designed for interpretable and robust feature selection.
Our system orchestrates a deliberative “debate” among multiple LLM agents, each operating within a defined role, allowing them to collectively evaluate feature relevance and provide detailed justifications for their selections. 
We conducted an empirical evaluation in the cybersecurity domain, focusing on an IoT intrusion detection use case with the CIC-DIAD 2024 dataset. The comparative analysis rigorously evaluates LLM-FS-Agent against prominent baselines, including LLM-Select and traditional methods such as PCA, across different feature subset sizes.
The results show that LLM-FS-Agent consistently achieves superior or comparable performance while substantially reducing downstream classifier training time, with an average reduction of 46\% (statistically significant $0.094 s$, $p = 0.028$ for XGBoost).
These findings confirm that the deliberative architecture not only enhances decision-making transparency but also improves computational efficiency, underscoring its potential as a reliable and practical solution for real-world applications.
\end{abstract}

\section{Introduction}\label{sec:introduction}

In the contemporary landscape of digital transformation, machine learning (ML) pipelines function as critical engines of innovation across diverse scientific and industrial domains. Their effectiveness, however, is often constrained by the “curse of dimensionality” \cite{verleysen2005curse}, a phenomenon in which the exponential growth of both data volume and feature count not only escalates computational demands but also jeopardizes model performance by increasing the risk of overfitting and limiting generalization capabilities \cite{domingos2012few}. Consequently, addressing this challenge is essential for the robust and reliable deployment of ML systems.

To mitigate the challenges associated with high-dimensional data, feature selection (FS) has emerged as a pivotal strategy within dimensionality reduction (DR) techniques, seeking to identify and retain only the most relevant original features while preserving their semantic integrity.
Traditional FS methods are generally categorized into three main types, filter, wrapper, and embedded approaches \cite{lazar2012survey, guyon2003introduction, chandrashekar2014survey}. Filter methods, such as mutual information and the Fisher score, rank features based on statistical criteria independent of downstream models \cite{lewis1992feature, ding2005minimum, gu2011generalized}. Wrapper methods, including sequential selection and Recursive Feature Elimination (RFE), perform heuristic searches to identify feature subsets that optimize model performance \cite{kohavi1997wrappers, guyon2002gene}. Embedded methods, such as Lasso, incorporate feature selection directly into model training through regularization mechanisms that promote sparsity \cite{tibshirani1996regression, yuan2006model}.
While these approaches are effective at reducing data complexity, a fundamental limitation persists, they often operate as “black boxes”, offering limited insight into the rationale behind the selection or exclusion of specific features. This lack of interpretability can undermine trust and accountability, particularly in sensitive domains such as cybersecurity, where transparency is critical for reliable threat analysis and decision-making.

\subsection{LLMs for Feature Selection}
In recent years, the advanced reasoning capabilities of Large Language Models (LLMs) have opened a new frontier for FS. As highlighted by Li et al. \cite{li2024exploring}, LLM-based FS methods can be categorized into two distinct approaches namely Text-based and Data-driven FS.

\textbf{Text-based FS} approaches leverage the extensive prior knowledge of LLMs to perform semantic associations using descriptive context, such as feature names and task descriptions, without requiring numerical data. This paradigm is particularly effective in low-resource settings and exhibits strong scaling behavior with model size.
A notable example is the LLM-Select framework \cite{jeong2024llm}, which introduces several prompting strategies, including LLM-SCORE, LLM-RANK, and LLM-SEQ. The authors demonstrate that sufficiently large LLMs can achieve performance competitive with traditional data-driven methods, even in the absence of access to the original training data; however, the effectiveness of this approach is strongly dependent on model scale.
Building on this concept, Retrieval-Augmented Feature Selection (RAFS) \cite{li2024exploring} extends text-based FS to domain-specific settings by incorporating external information as auxiliary context, demonstrating its practical utility in real-world applications medical application.

\textbf{Data-driven FS} approaches, in contrast, provide the LLM with specific data samples or values to perform statistical inference and uncover correlations \cite{li2024exploring}. The prompts in this paradigm include data points as few-shot examples, enabling LLMs to infer relationships and carry out basic statistical analyses.
While this method can be effective, it is often constrained by the limited context window of LLMs, which hampers their ability to process long sequences of data points as sample sizes increase, resulting in a notable decline in performance \cite{dong2023bamboo, liu2024lost}. Consequently, the effectiveness of data-driven FS with LLMs is restricted, often yielding lower performance than text-based approaches in low-resource settings and rendering it impractical for full-shot scenarios.

Other related works have explored more advanced applications of LLMs in feature selection. For instance, Han et al. \cite{han2024large} employed LLMs as "feature engineers" to generate meta-features that enhance the performance of downstream models. Similarly, Liu et al. \cite{liu2024ice} proposed the In-Context Evolutionary Search (ICE-SEARCH), which leverages LLMs to optimize selected features by filtering them based on test scores.

\subsection{Agentic and Multi-Agent LLM architectures}

The limitations of single-agent prompting have motivated the development of multi-agent LLM architectures as a powerful paradigm for solving complex, multi-step tasks by leveraging modular design, collective reasoning, and improved interpretability \cite{wang2024survey, xi2023rise}. 
These architectures provide the foundational basis for our deliberative approach. 
They extend LLM capabilities by integrating tools and APIs \cite{yang2024gpt4tools, schick2024toolformer}, enabling the execution of actions and the resolution of complex, multi-step problems. 
This paradigm has been explored in data science, where LLMs can leverage statistical tools and software to assist in data processing and analysis \cite{hong2024data, fang2024large}.
Building on this foundation, recent work has introduced multi-agent frameworks such as Tree-of-Thought (ToT) \cite{yao2023tree}, Graph-of-Thought (GoT) \cite{besta2024graph}, and Mixture-of-Agents (MoA) \cite{wang2024mixture}, demonstrating that orchestrating multiple agents can yield more reliable and robust outcomes compared with relying on a single model. These frameworks emphasize principles of agent coordination, committee-based decision-making, and structured deliberation, thereby offering promising avenues for generating well-founded justifications and enhancing interpretability.

While existing multi-agent systems are effective for general reasoning and coding tasks, they lack the specialized, domain-aware scrutiny required for feature selection. 
General deliberation frameworks do not inherently address critical data science considerations, such as balancing feature relevance with redundancy, mitigating collinearity risks, or defending against adversarial manipulation, which are particularly important in cybersecurity applications.

Addressing these limitations, this study investigates how LLMs can be leveraged within a structured, multi-agent architecture to achieve transparent, justifiable, and high-performing feature selection for high-dimensional data. We introduce LLM-FS-Agent, which assigns specialized roles such as a Refiner for statistical context and a Challenger for adversarial critique to facilitate structured debates around feature metadata and semantic utility, surpassing the simple "agree or disagree" loops of conventional reasoning frameworks. Unlike single-agent prompting and traditional black-box methods, LLM-FS-Agent enables a deliberative and auditable feature selection process, producing human-interpretable rationales that ensure transparency, robustness, and superior generalizable performance.

To validate the practical relevance and robustness of the proposed methodology, we conducted an empirical evaluation in the context of an Intrusion Detection System (IDS) for IoT devices. The assessment focuses on network traffic classification using the CIC-DIAD 2024 dataset \cite{cicdiad2024}, and includes a direct comparison between LLM-FS-Agent and the LLM-Select method. This analysis highlights the improvements in interpretability, consistency, and predictive performance enabled by the multi-agent architecture.
In summary, the main contributions of this work are:
\begin{enumerate}
    \item A transparent feature selection approach in which role-specialized LLM agents (Initiator, Refiner, Challenger, and Judge) engage in a structured deliberation process.
    
    \item A comprehensive qualitative transparency evaluation of the deliberation process, including confidence scores and detailed justifications for all selected features.

    \item An empirical evaluation on an intrusion detection task, comparing the predictive performance of LLM-FS-Agent against single-agent approach across diverse downstream classifiers (XGBoost, RF, SVC, LR) and varying feature subset sizes.
\end{enumerate}

The paper is organized as follows. Section 2 describes the experimental methodology and architectural specifications of LLM-FS-Agent. Section 3 presents the empirical findings, followed by a critical discussion and promising directions for future research in Section 4. The final section summarizes the main contributions.

\section{Methods}\label{sec:Methods}

\begin{figure}[htbp]
    \centering
    \includegraphics[width=\linewidth]{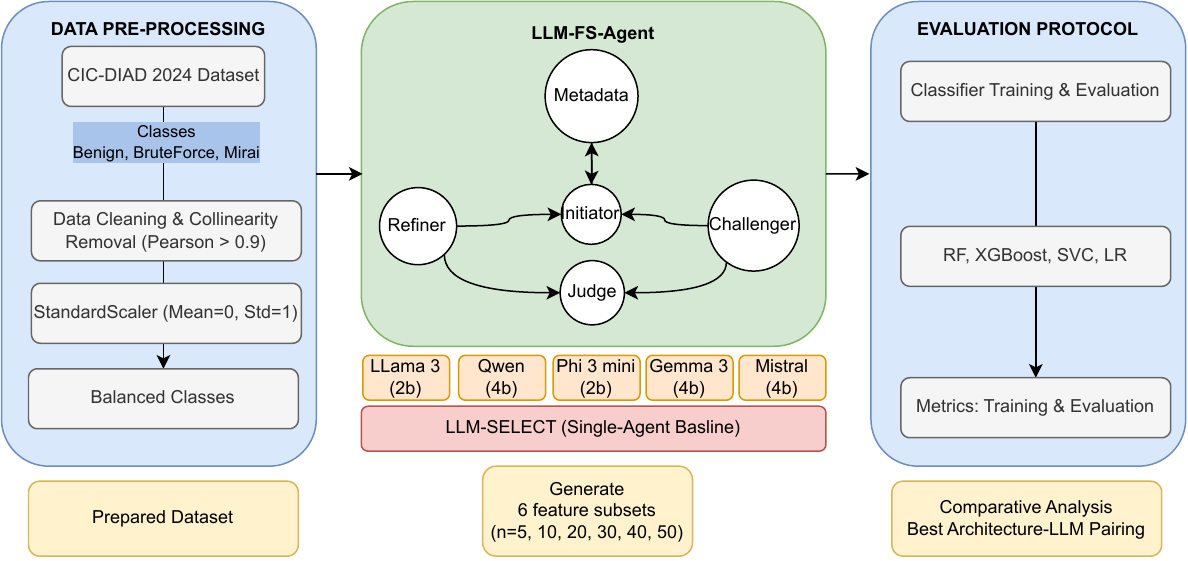} 
    \caption{Pipeline and evaluation protocol of LLM-FS-Agent}
    \label{fig:eval-proto}
\end{figure}

This section presents the experimental design, data processing pipeline, evaluation protocol, and architecture of LLM-FS-Agent. As illustrated in Figure \ref{fig:eval-proto}, this methodology facilitates a rigorous and transparent comparison with an LLM-based feature selection method from prior literature.

\subsection{Experimental setup and data preprocessing}

Experiments were conducted on the CIC-DIAD 2024 dataset \cite{cicdiad2024} specifically designed for IoT cybersecurity intrusion detection. The dataset comprises 84 features in total. For this study, we selected a representative subset comprising samples from three categories, Benign, Brute Force, and Mirai. The data were subjected to a sequence of preprocessing steps to ensure quality and consistency across all experiments (Table \ref{tab:dataset-before-after}).
This included initial data cleaning and the removal of features exhibiting high collinearity defined by a Pearson correlation coefficient greater than 0.9. The remaining numerical features were then standardized using StandardScaler to a mean of 0 and a standard deviation of 1. 
To mitigate the class imbalance present in the original dataset, random undersampling was applied to achieve a balanced distribution of samples across the selected categories.

\begin{table}[htbp]
    \caption{Dataset distribution before and after preprocessing}
    \label{tab:dataset-before-after}
    \centering
    \begin{tabular}{lrr|rr}
    \toprule
    & \multicolumn{2}{c}{\textbf{Original}} 
    & \multicolumn{2}{c}{\textbf{Pre-processed}} \\
    \cmidrule(lr){2-3} \cmidrule(lr){4-5}
    \textbf{Class} & \textbf{Count} & \textbf{\%} 
                   & \textbf{Count} & \textbf{\%} \\
    \midrule
    Benign      & 183,595 & 94.0 & 3,619  & 29.2 \\
    Mirai       & 5,170   & 2.6  & 5,170  & 41.7 \\
    BruteForce  & 3,619   & 1.9  & 3,619  & 29.2 \\
    \midrule
    \textbf{Total} & \textbf{192,384} & \textbf{100.0} 
                   & \textbf{12,408}  & \textbf{100.0} \\
    \bottomrule
    \end{tabular}
\end{table}

\subsection{Evaluation Protocol}

To assess the efficacy and generalizability of the selected feature subsets, a comprehensive evaluation protocol was employed. The subsets were used to train four widely adopted classical ML classifiers namely Random Forest (RF), XGBoost, Support Vector Classifier (SVC), and Logistic Regression (LR). These classifiers were chosen because they represent a diverse set of model types (ensemble, boosting, kernel-based, and linear) thereby providing a robust assessment of the generalizability of the selected features.

The performance of the classifiers was evaluated using two primary metrics, accuracy and the area under the receiver operating characteristic curve (AUC). 
To ensure a fair comparison between LLM-FS-Agent and LLM-Select, the same LLMs were deployed locally via the Ollama framework, including Llama 3.2 (2B), Gemma\_3 (4B), Qwen (4B), Phi-3 Mini (3B), and Mistral (4B).
For each method and corresponding LLM, the features were first ranked according to their importance scores. Based on this ranking, six separate feature subsets were constructed by selecting the top $n$ features ($n \in {5, 10, 20, 30, 40, 50}$). For example, the first subset contained only the top 5 ranked features, the second contained the top 10, and so on. Each of these subsets was then used independently to train the four downstream classifiers, enabling a systematic evaluation of feature subset size on model performance across all experimental variables, including the feature selection method and subset size, the underlying language model, the downstream classifier, and the chosen performance metrics.

\subsection{LLM-FS-Agent: Architecture design and the Deliberative Feature Selection process}

The proposed LLM-FS-Agent is designed to enable transparent feature selection. Its architecture is built around a collaborative debate mechanism among multiple LLM agents, aiming to overcome the black-box limitations of both conventional techniques and direct LLM prompting approaches.
As illustrated in Figure~\ref{fig:llm-fs-agent-architecture}, the system receives as input the feature names along with a textual description of the prediction task. This information is processed through a sequence of LLM agents, each assigned a specific role in constructing a comprehensive argument either supporting or contesting the importance of each feature. The roles of these agents are defined as follows:

\begin{enumerate}
    \item \textbf{The Initiator Agent}: Conducts an initial semantic analysis of each feature based on the task description, providing a preliminary relevance assessment.

    \item \textbf{The Refiner Agent}: Enhances the Initiator’s analysis by generating supporting arguments, including metadata such as the mean and standard deviation of the feature–target correlation.
    
    \item \textbf{The Challenger Agent}: Critically examines the Initiator’s arguments to identify weaknesses, redundancies, or biases, providing counter-arguments in a structured, peer-review–like manner.
    
    \item \textbf{The Judge Agent}: Acts as the final arbiter by synthesizing all arguments and counter-arguments. It assigns a final importance score ($S_{\text{final}}$) to each feature using a weighted combination of the refined ($S_{\text{refined}}$) and challenged ($S_{\text{challenged}}$) scores, described in Equation~\ref{eq:final_score}.
    
    \begin{equation}
    \label{eq:final_score}
        \begin{cases}
        S_{\text{final}} = w_r \cdot S_{\text{refined}} + w_c \cdot S_{\text{challenged}} \\
        w_r + w_c = 1
        \end{cases}
    \end{equation}
    
    \vspace{0.2cm}

    \begin{minipage}{\linewidth}
        \centering
        \includegraphics[width=0.58\linewidth]{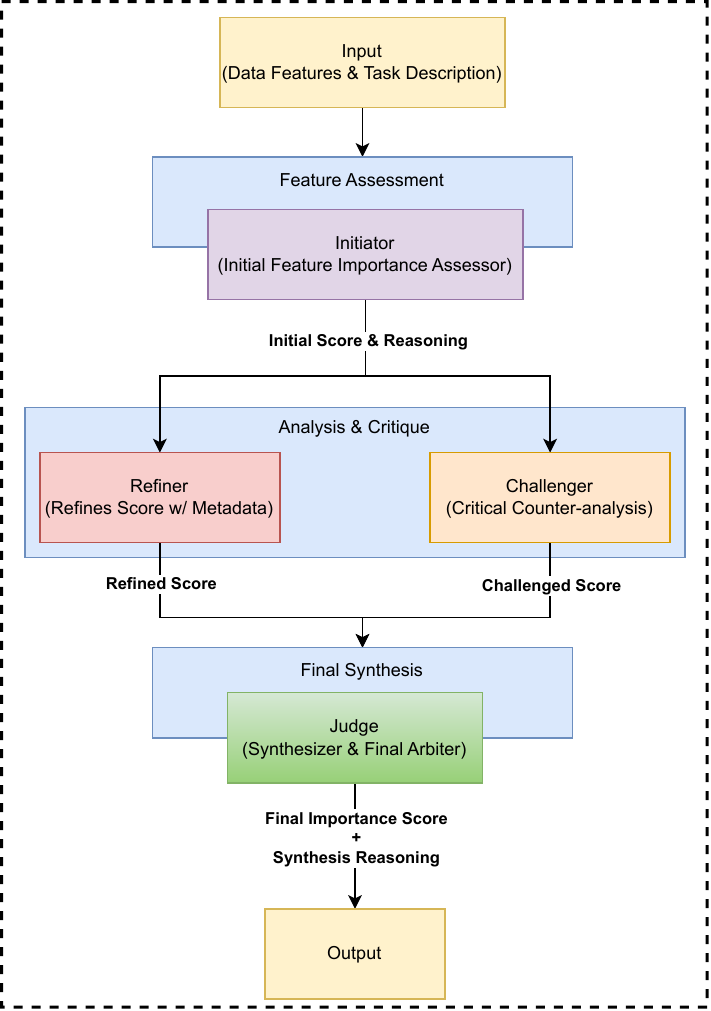}
        \captionof{figure}{Architecture of the LLM-FS-Agent}
        \label{fig:llm-fs-agent-architecture}
    \end{minipage}
\end{enumerate}

\begin{figure}[htbp]
    \centering
    \includegraphics[scale=.65]{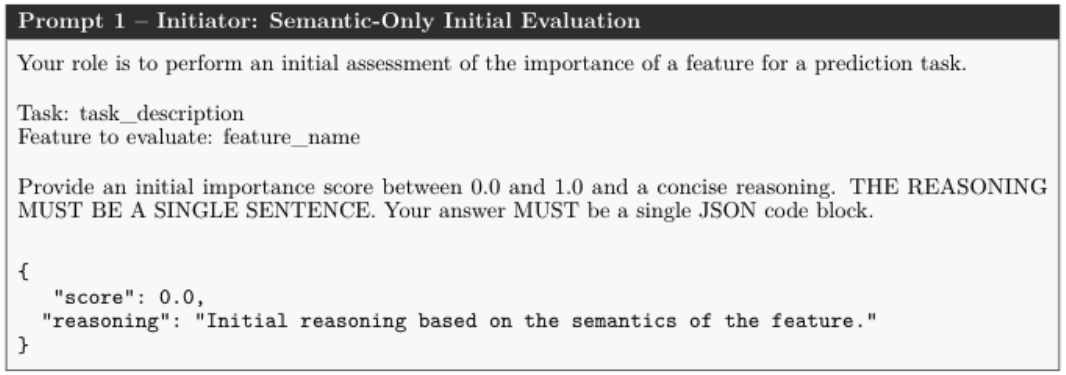}
    \caption{The initiator LLM prompt template}
    \label{fig:prompt-initiator}
\end{figure}

\begin{figure}[htbp]
    \centering
    \includegraphics[scale=.65]{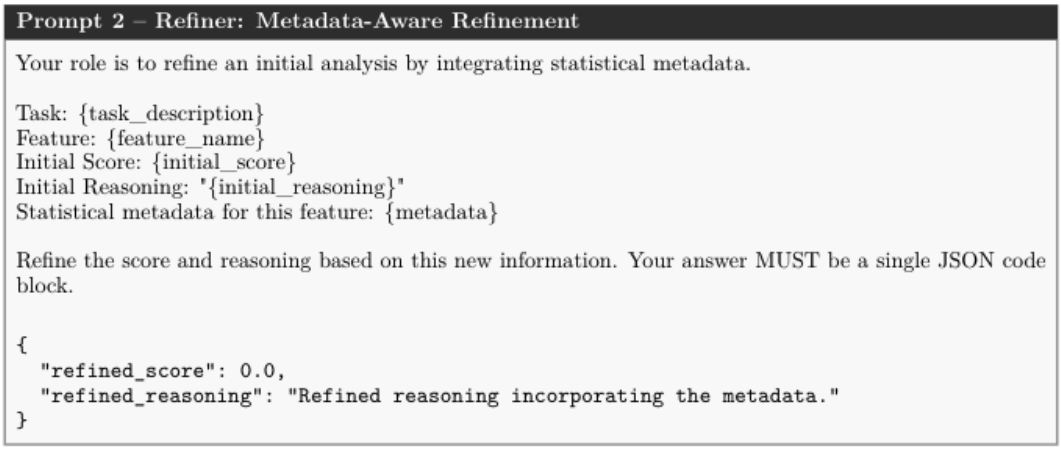}
    \caption{The refiner LLM prompt template}
    \label{fig:prompt-refiner}
\end{figure}

\begin{figure}[htbp]
    \centering
    \includegraphics[scale=.65]{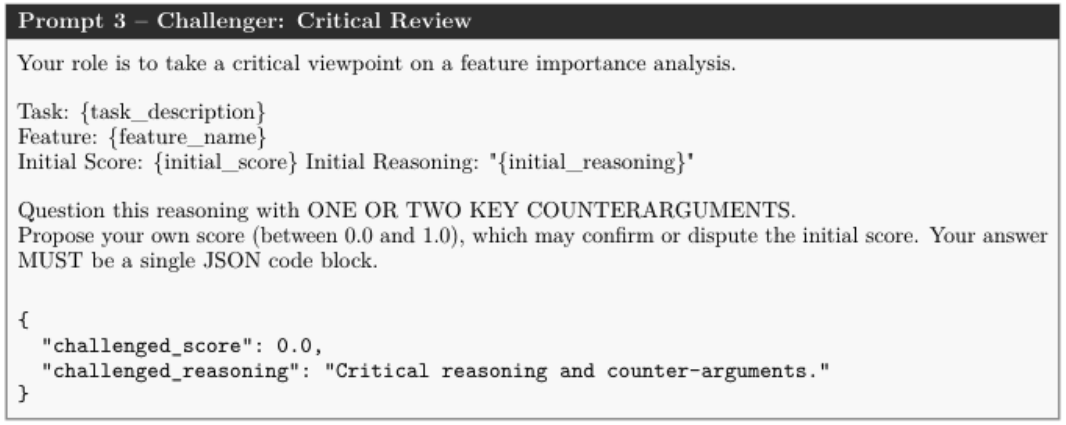}
    \caption{The challenger LLM prompt template}
    \label{fig:prompt-challenger}
\end{figure}

\begin{figure}[htbp]
    \centering
    \includegraphics[scale=.65]{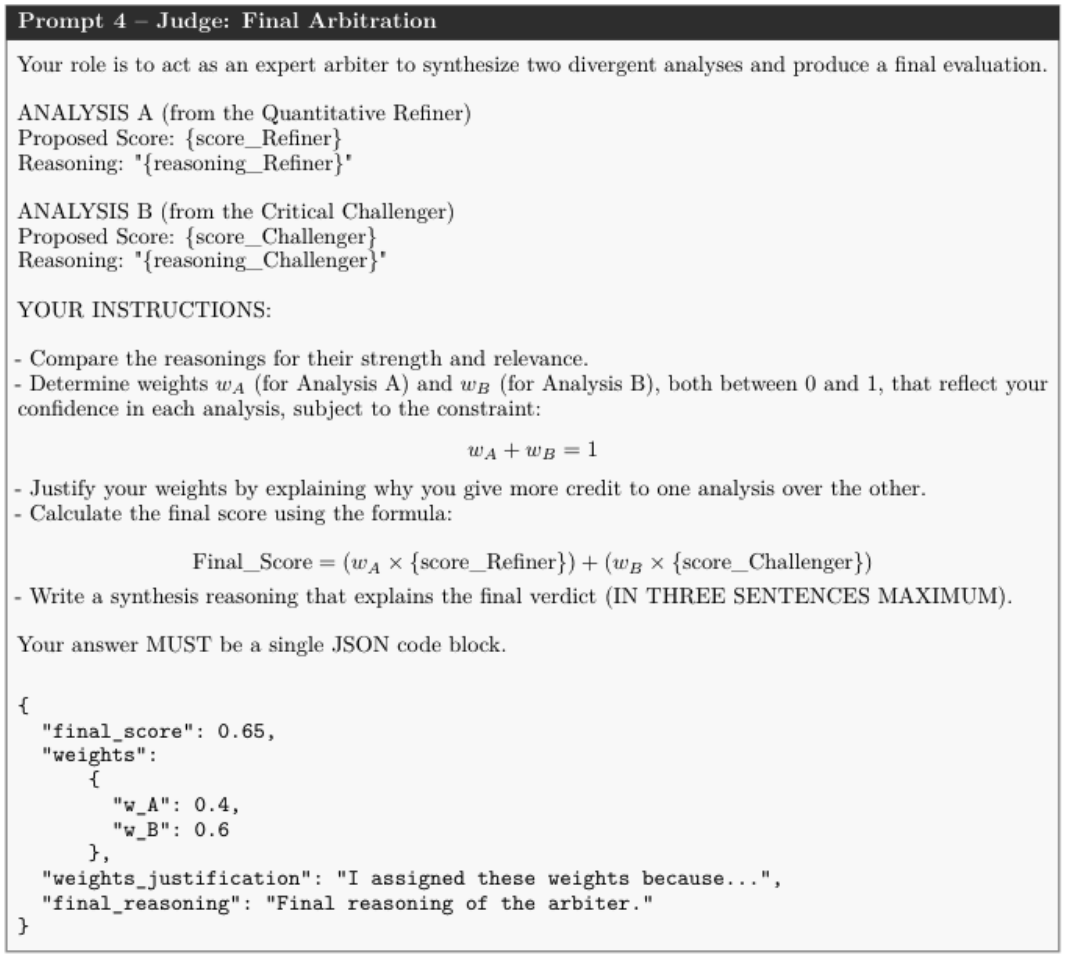}
    \caption{The judge LLM prompt template}
    \label{fig:prompt-judge}
\end{figure}

To ensure full transparency and reproducibility, the exact prompts used for each agent within the LLM-FS-Agent architecture are presented in Figures~\ref{fig:prompt-initiator}, \ref{fig:prompt-refiner}, \ref{fig:prompt-challenger}, and \ref{fig:prompt-judge}. 
These prompts define the specific instructions, contextual information, and expected output format for each stage of the deliberative process.

\section{Results}\label{sec:Results}

This evaluation examines the performance of four classifiers across six predefined feature subset sizes, with particular emphasis on predictive accuracy and the interpretability of the deliberative process.

\subsection{Quantitative performance and generalizability}\label{subsec:quant_results}

\begin{figure}[htbp]
    \centering

    \begin{minipage}[t]{0.48\linewidth}
        \centering
        \includegraphics[height=8cm,width=7cm,trim=0 0 150 42,clip]{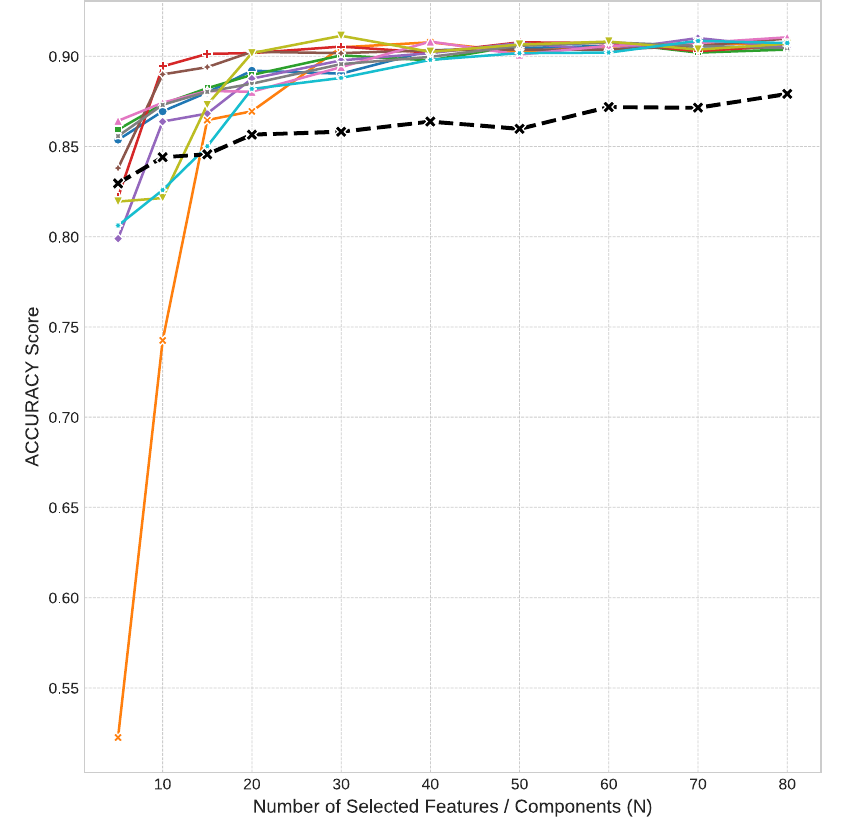}
        \caption*{XGBoost}
    \end{minipage}
    \begin{minipage}[t]{0.48\linewidth}
        \centering
        \includegraphics[height=8cm,width=9cm,trim=20 0 0 41,clip]{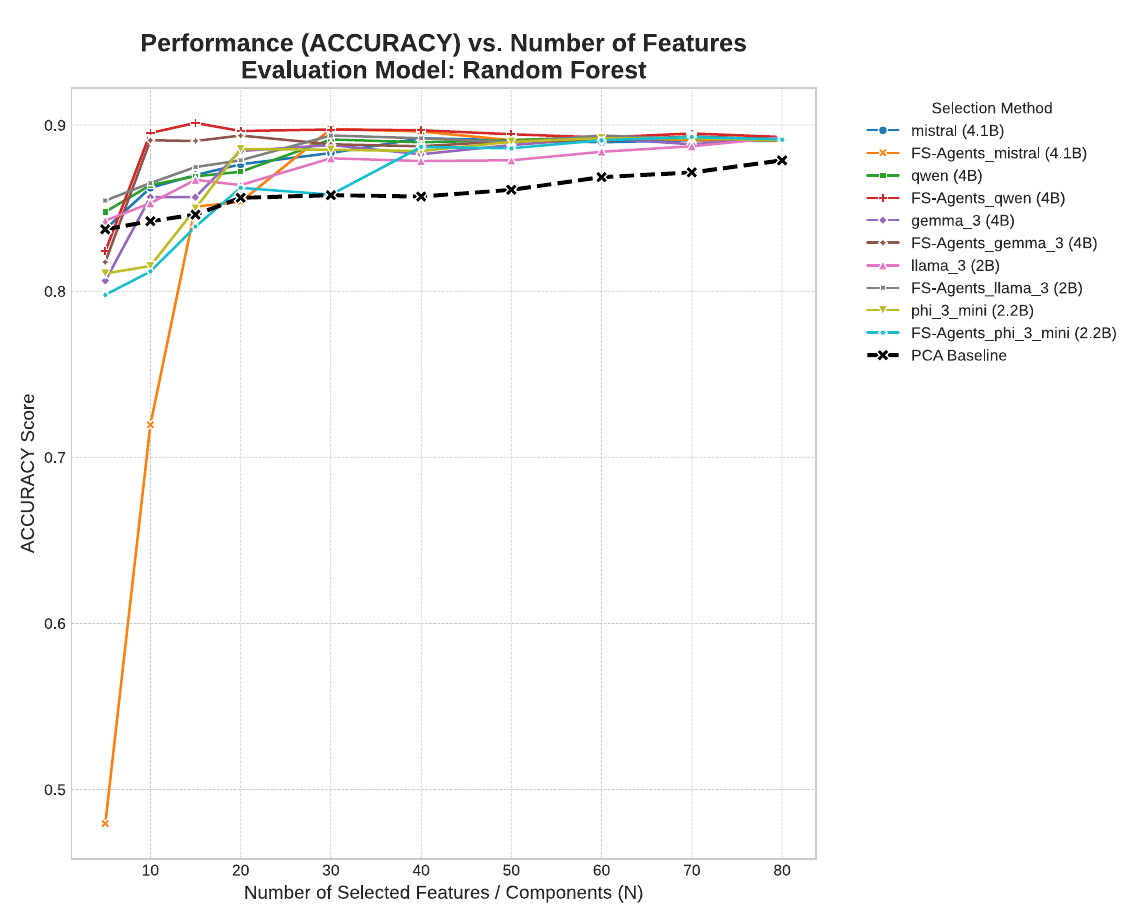}
        \caption*{Random Forest}
    \end{minipage}

    \vspace{0.5cm}  

    \begin{minipage}[t]{0.48\linewidth}
        \centering
        \includegraphics[height=8.15cm,width=7cm,trim=0 0 150 48,clip]{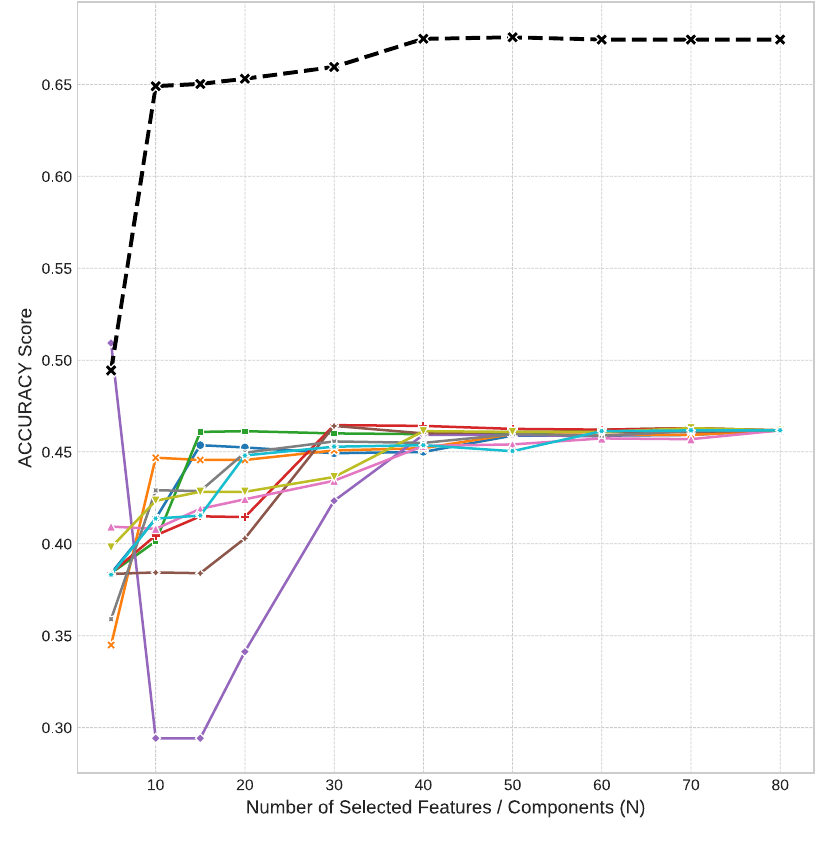}
        \caption*{SVC}
    \end{minipage}
    \begin{minipage}[t]{0.48\linewidth}
        \centering
        \includegraphics[height=8.15cm,width=9cm,trim=20 0 0 40,clip]{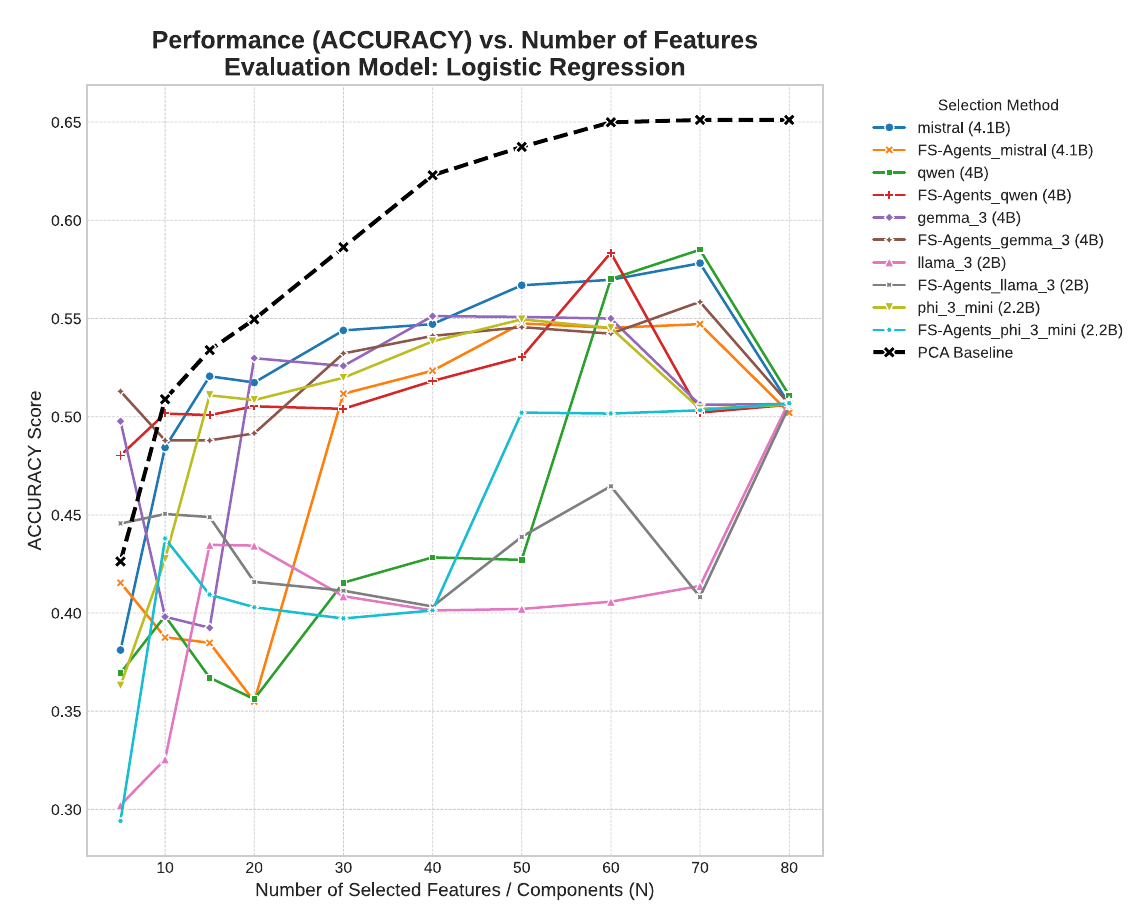}
        \caption*{Logistic Regression}
    \end{minipage}

    \caption{Performance curves comparing model accuracy for different classifiers}
    \label{fig:performance_curves}
\end{figure}

To evaluate LLM-based feature selection, we assessed downstream classifier performance using the selected feature subsets. As illustrated in Figure~\ref{fig:performance_curves}, XGBoost and RF consistently achieved high scores, demonstrating the robustness of LLM-based approaches compared to traditional methods such as PCA.

\begin{figure}[htbp]
    \centering
    \begin{minipage}{\textwidth}
        \centering
        \begin{minipage}[t]{0.95\textwidth} 
            \centering
            \begin{subfigure}[b]{0.32\textwidth}
                \centering
                \includegraphics[width=\linewidth]{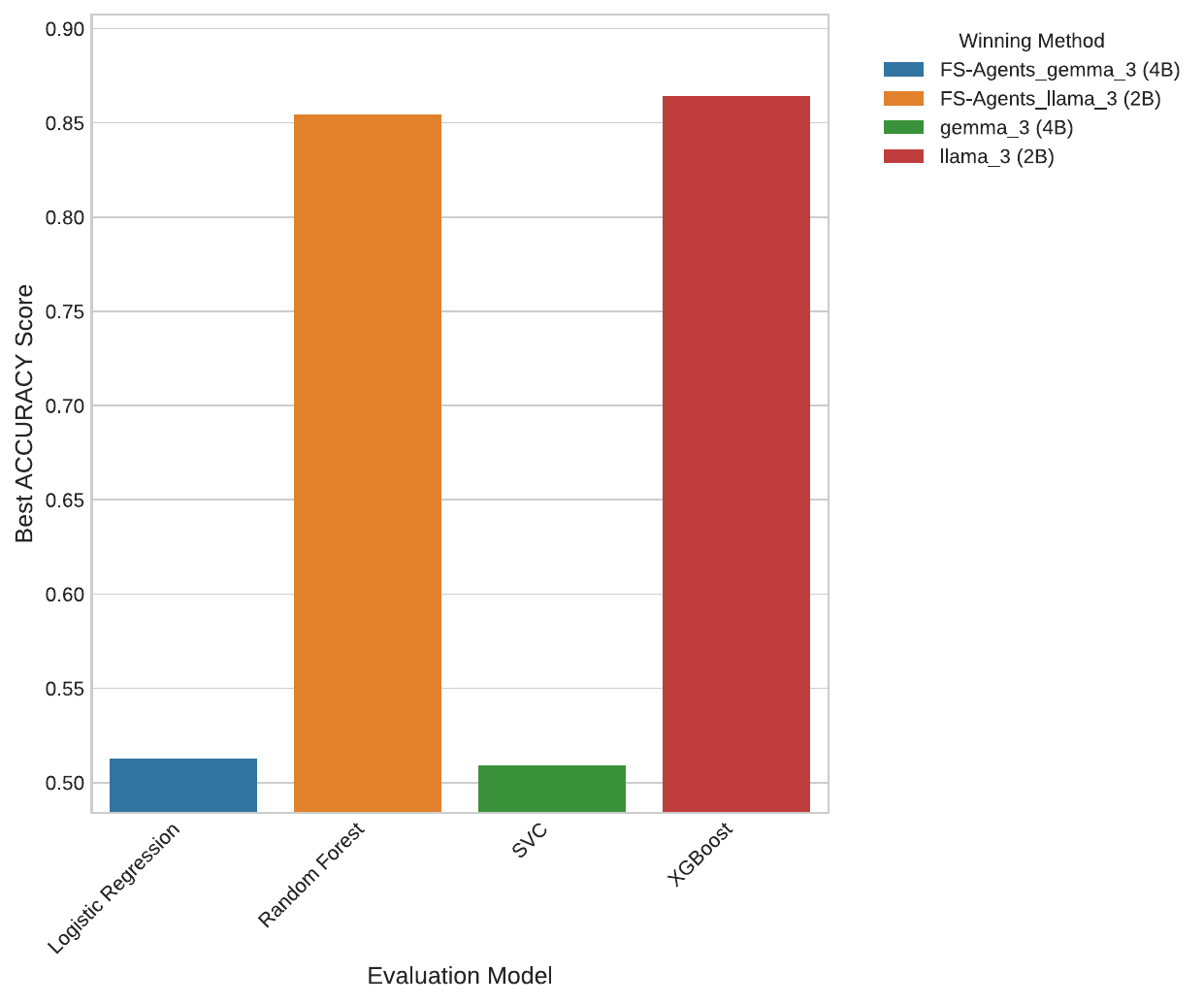}
                \caption{$n=5$} 
                \label{fig:acc_n5}
            \end{subfigure}
            \hfill
            \begin{subfigure}[b]{0.32\textwidth}
                \centering
                \includegraphics[width=\linewidth]{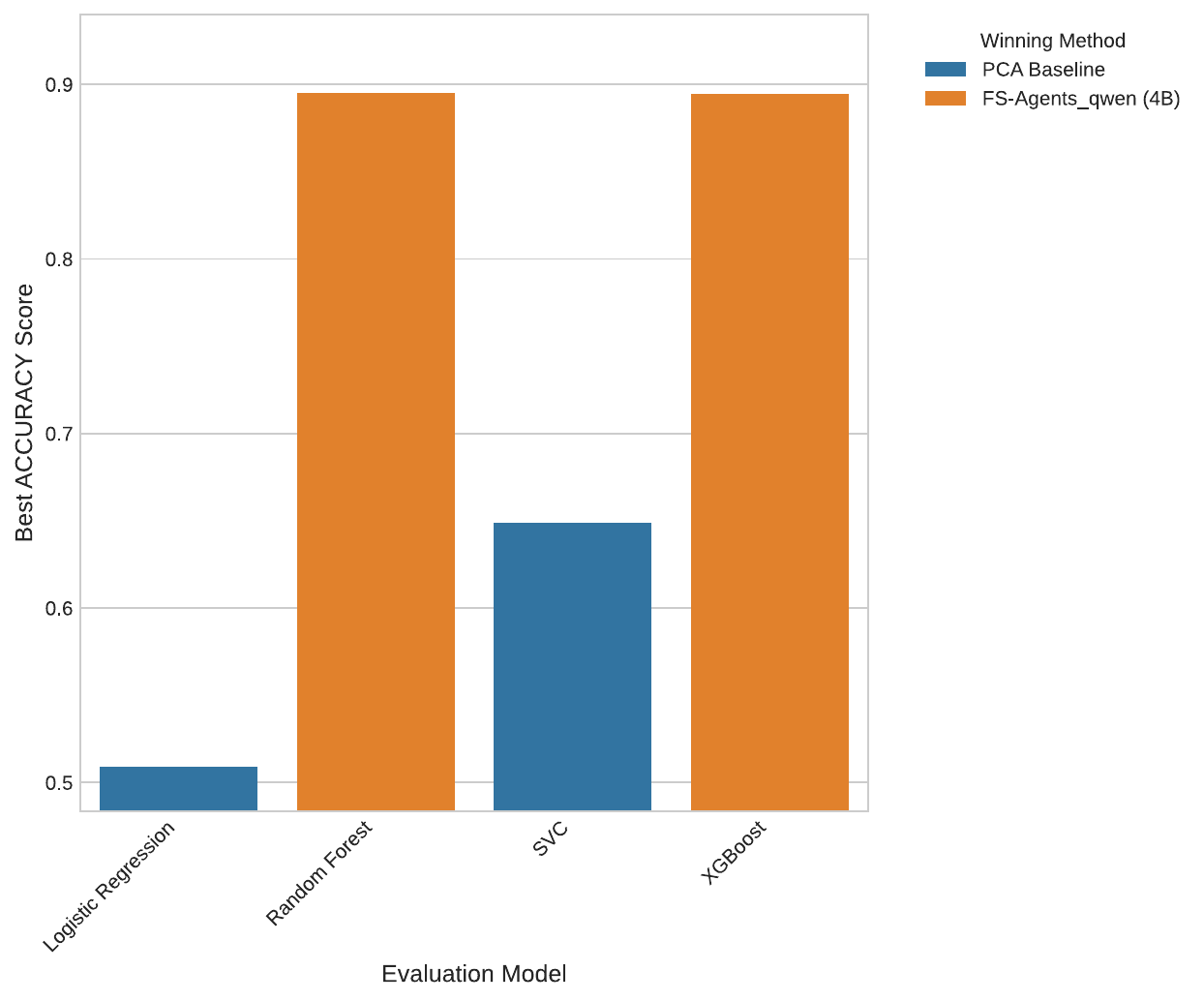}
                \caption{$n=10$}
                \label{fig:acc_n10}
            \end{subfigure}
            \hfill
            \begin{subfigure}[b]{0.32\textwidth}
                \centering
                \includegraphics[width=\linewidth]{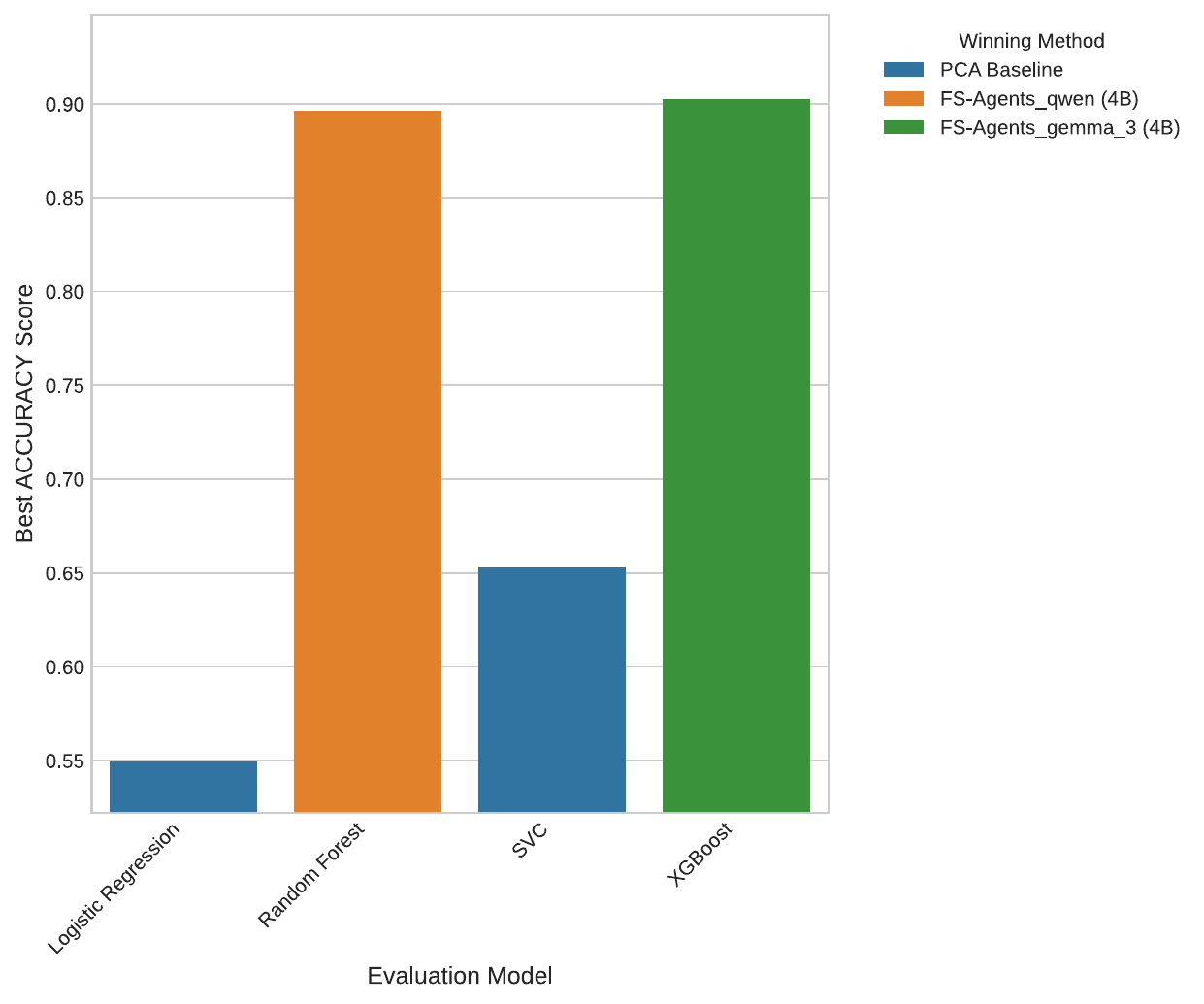}
                \caption{$n=20$}
                \label{fig:acc_n20}
            \end{subfigure}

            \vspace{0.5cm}

            \begin{subfigure}[b]{0.32\textwidth}
                \centering
                \includegraphics[width=\linewidth]{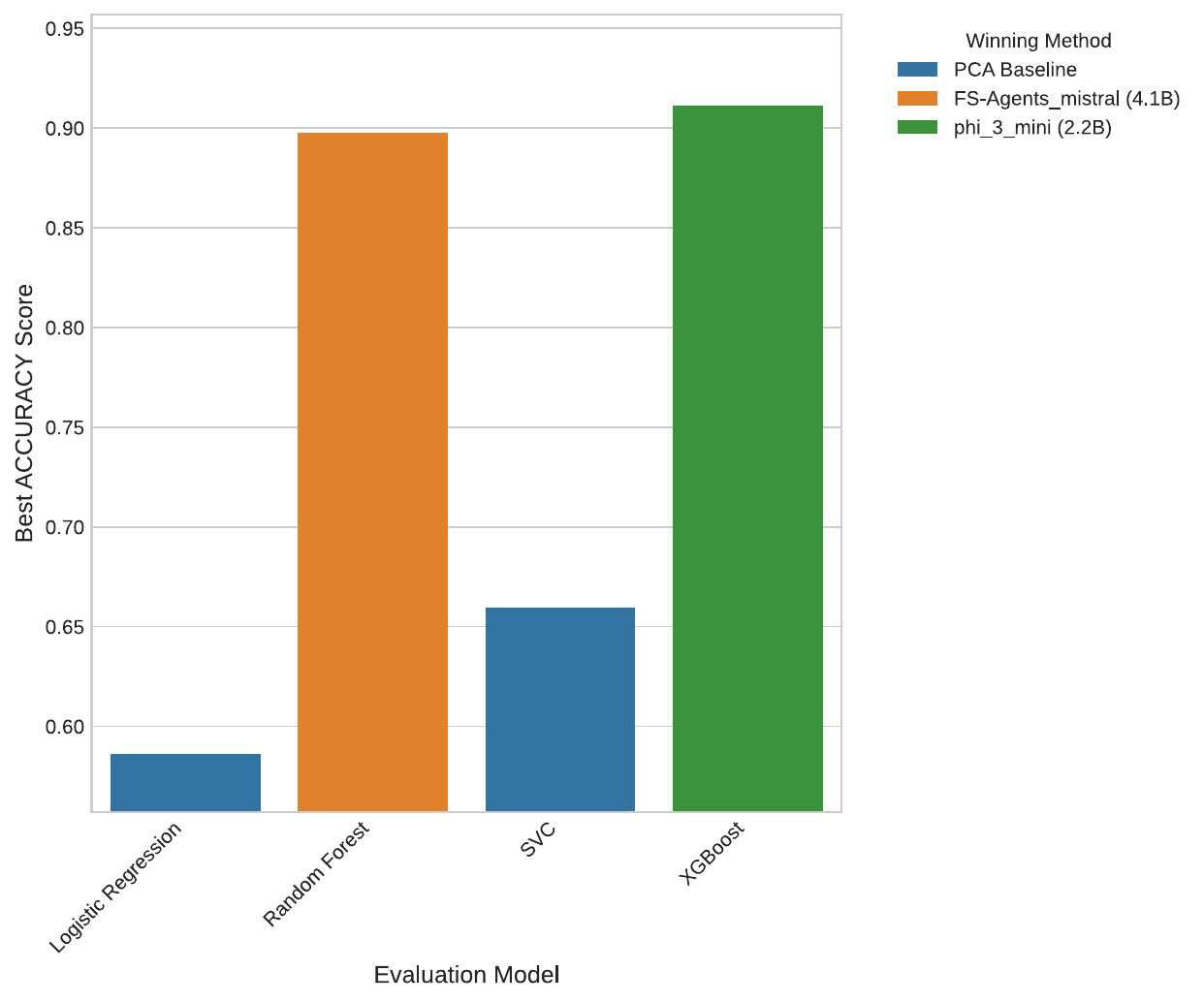}
                \caption{$n=30$}
                \label{fig:acc_n30}
            \end{subfigure}
            \hfill
            \begin{subfigure}[b]{0.32\textwidth}
                \centering
                \includegraphics[width=\linewidth]{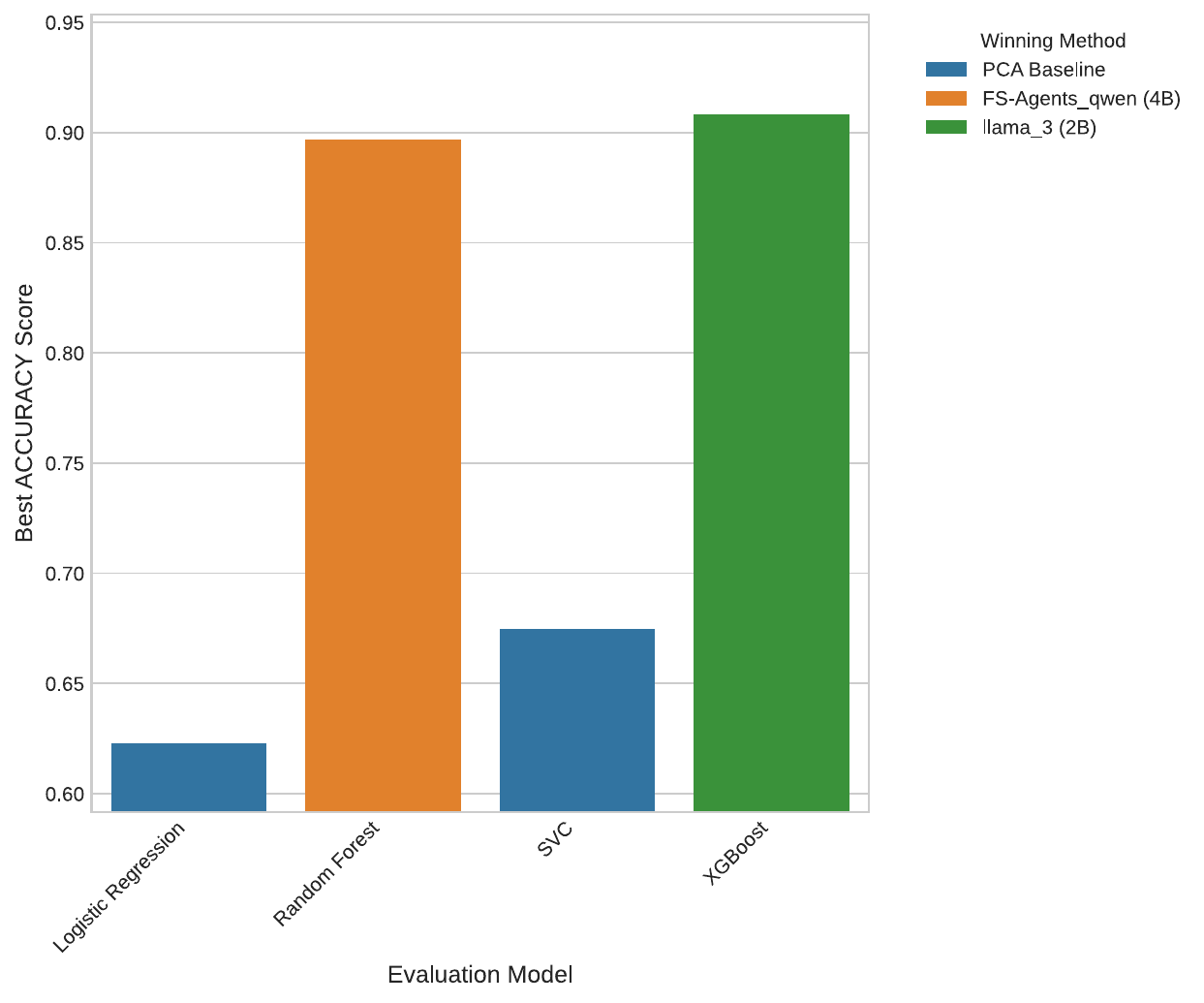}
                \caption{$n=40$}
                \label{fig:acc_n40}
            \end{subfigure}
            \hfill
            \begin{subfigure}[b]{0.32\textwidth}
                \centering
                \includegraphics[width=\linewidth]{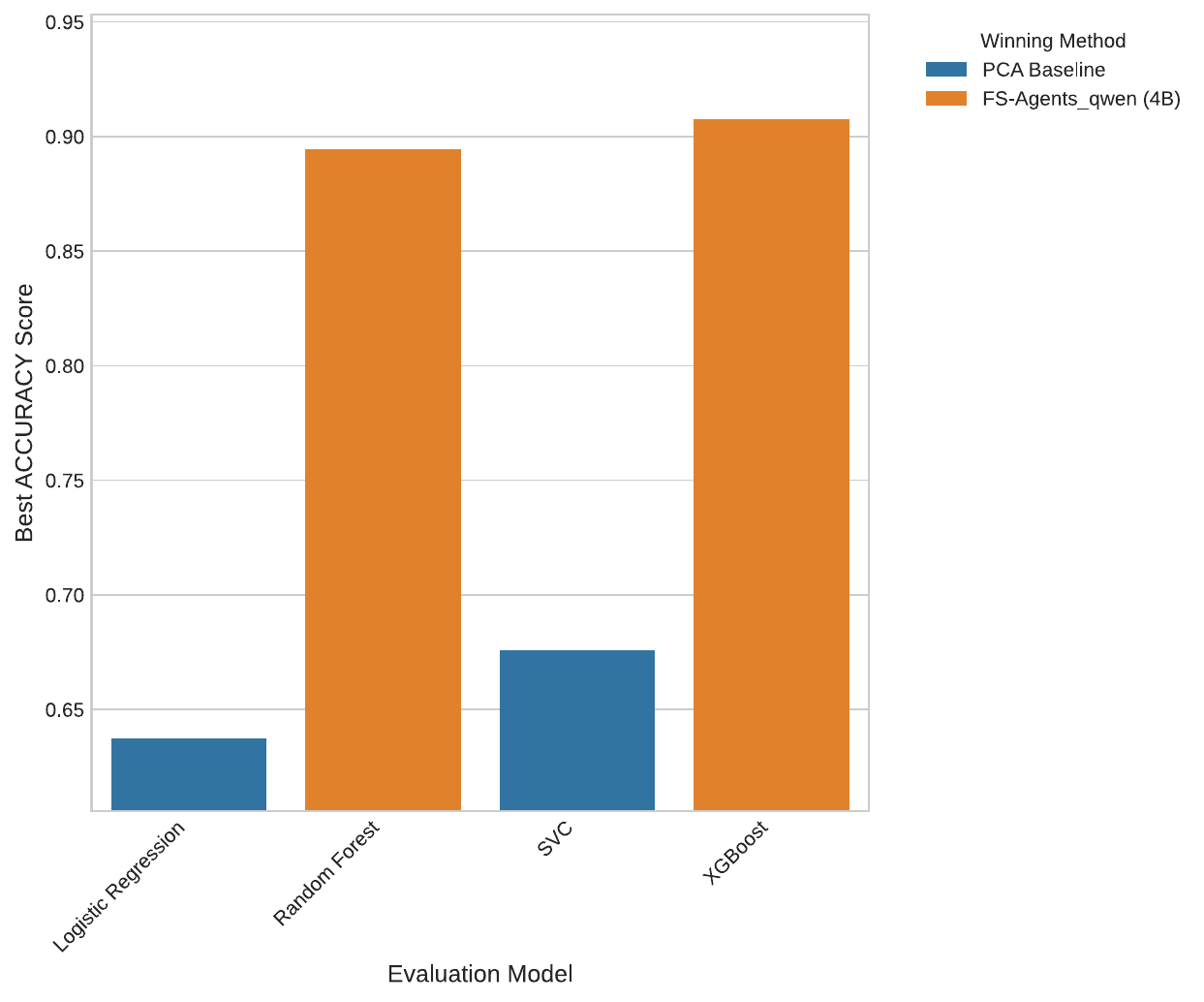}
                \caption{$n=50$}
                \label{fig:acc_n50}
            \end{subfigure}
        \end{minipage}
        \begin{minipage}[c]{0.04\textwidth}
            \centering
            \rotatebox{90}{\textbf{Accuracy}}
        \end{minipage}
    \end{minipage}

    \vspace{1cm}

    \begin{minipage}{\textwidth}
        \centering
       
        \begin{minipage}[t]{0.95\textwidth}
            \begin{subfigure}[b]{0.32\textwidth}
                \centering
                \includegraphics[width=\linewidth]{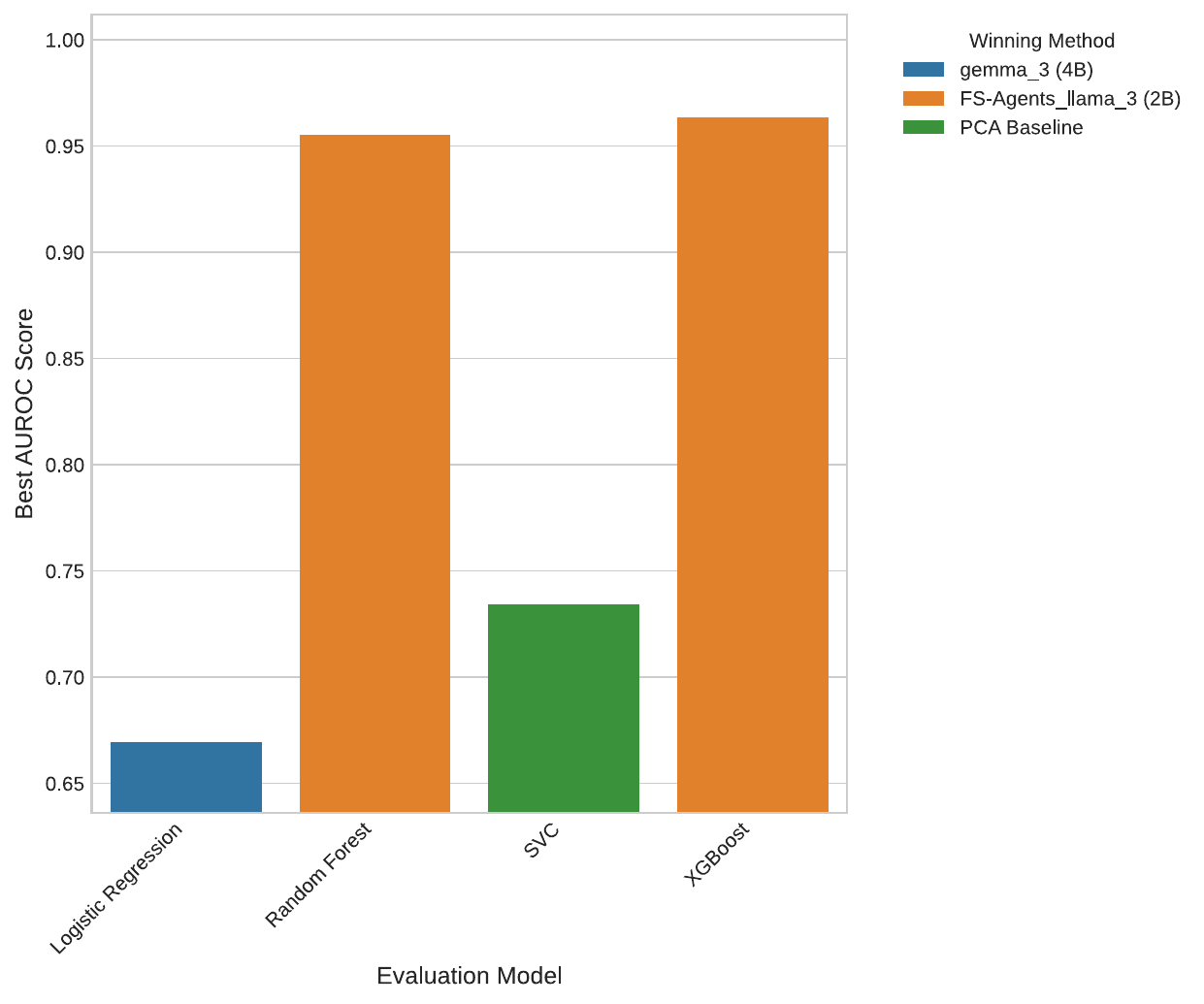}
                \caption{$n=5$} 
                \label{fig:auc_n5}
            \end{subfigure}
            \hfill
            \begin{subfigure}[b]{0.32\textwidth}
                \centering
                \includegraphics[width=\linewidth]{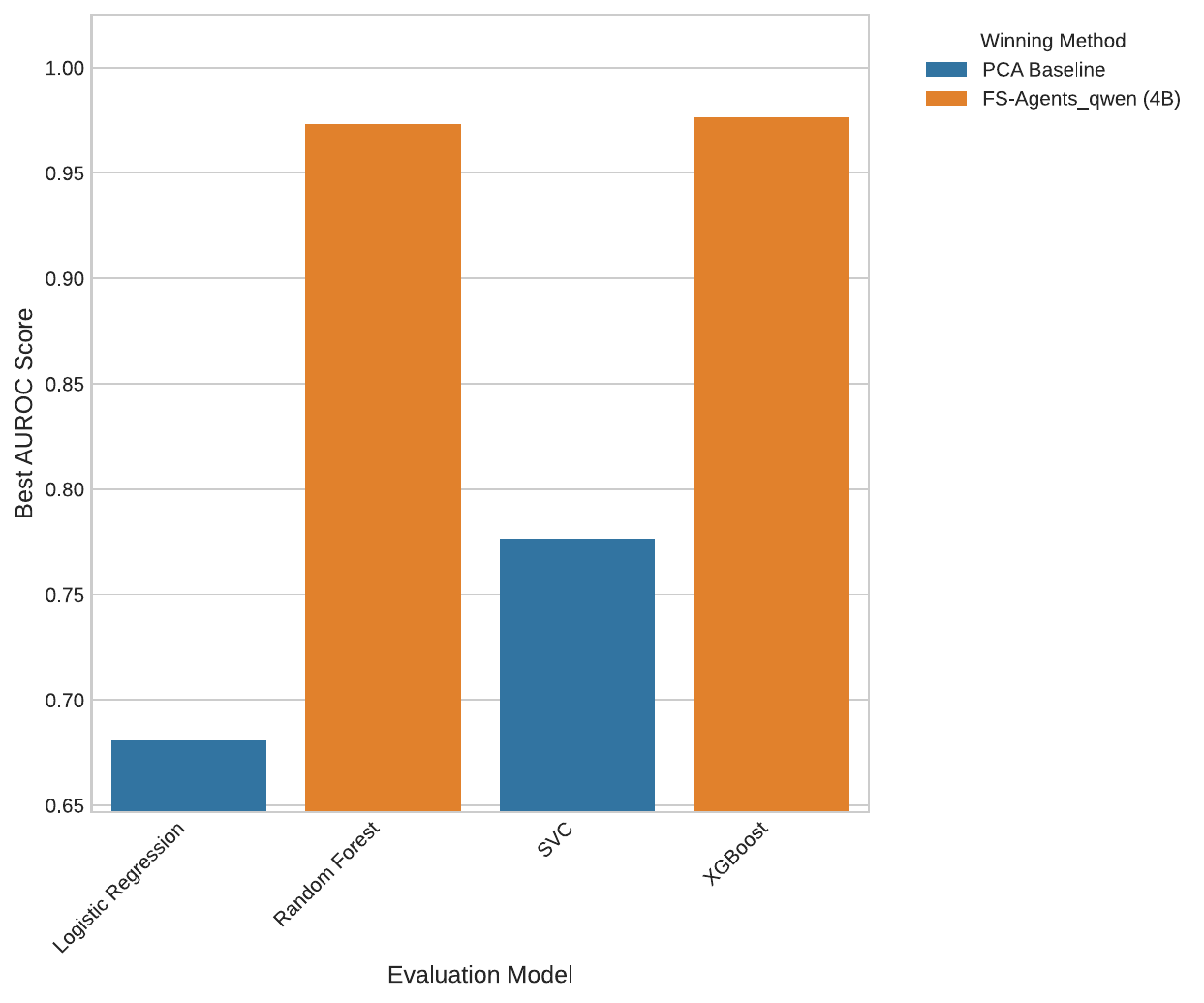}
                \caption{$n=10$}
                \label{fig:auc_n10}
            \end{subfigure}
            \hfill
            \begin{subfigure}[b]{0.32\textwidth}
                \centering
                \includegraphics[width=\linewidth]{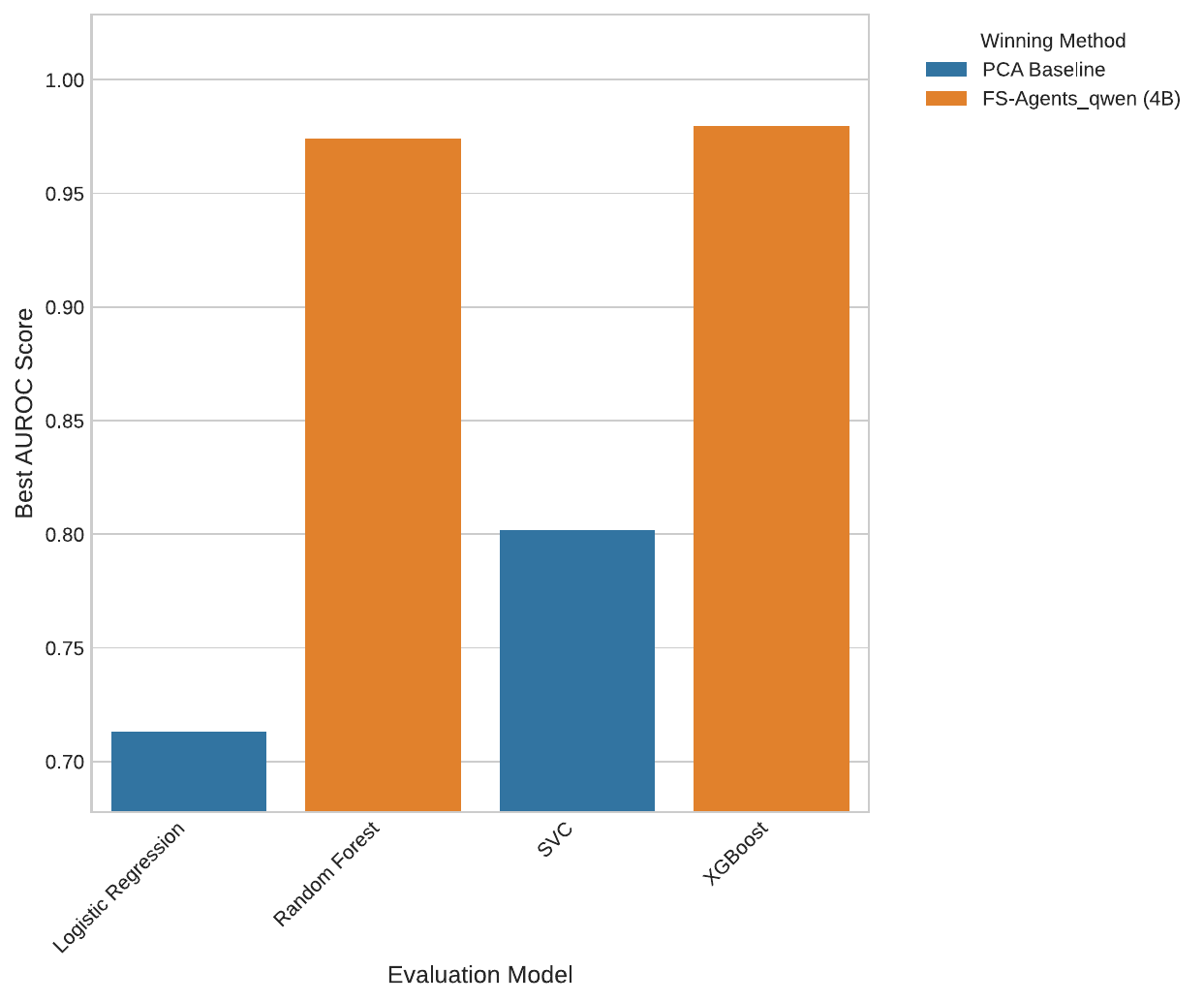}
                \caption{$n=20$}
                \label{fig:auc_n20}
            \end{subfigure}

            \vspace{0.5cm} 

            \begin{subfigure}[b]{0.32\textwidth}
                \centering
                \includegraphics[width=\linewidth]{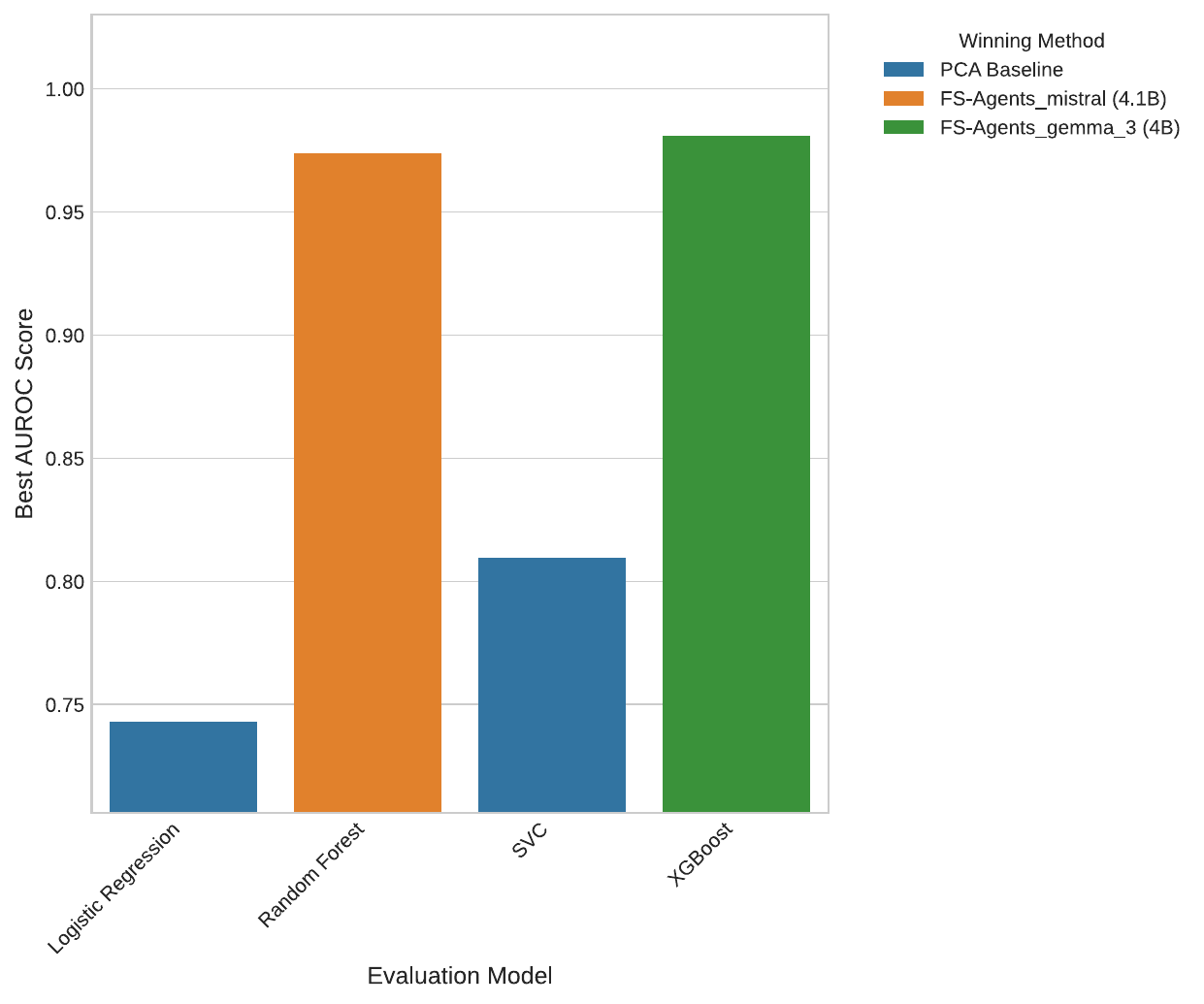}
                \caption{$n=30$}
                \label{fig:auc_n30}
            \end{subfigure}
            \hfill
            \begin{subfigure}[b]{0.32\textwidth}
                \centering
                \includegraphics[width=\linewidth]{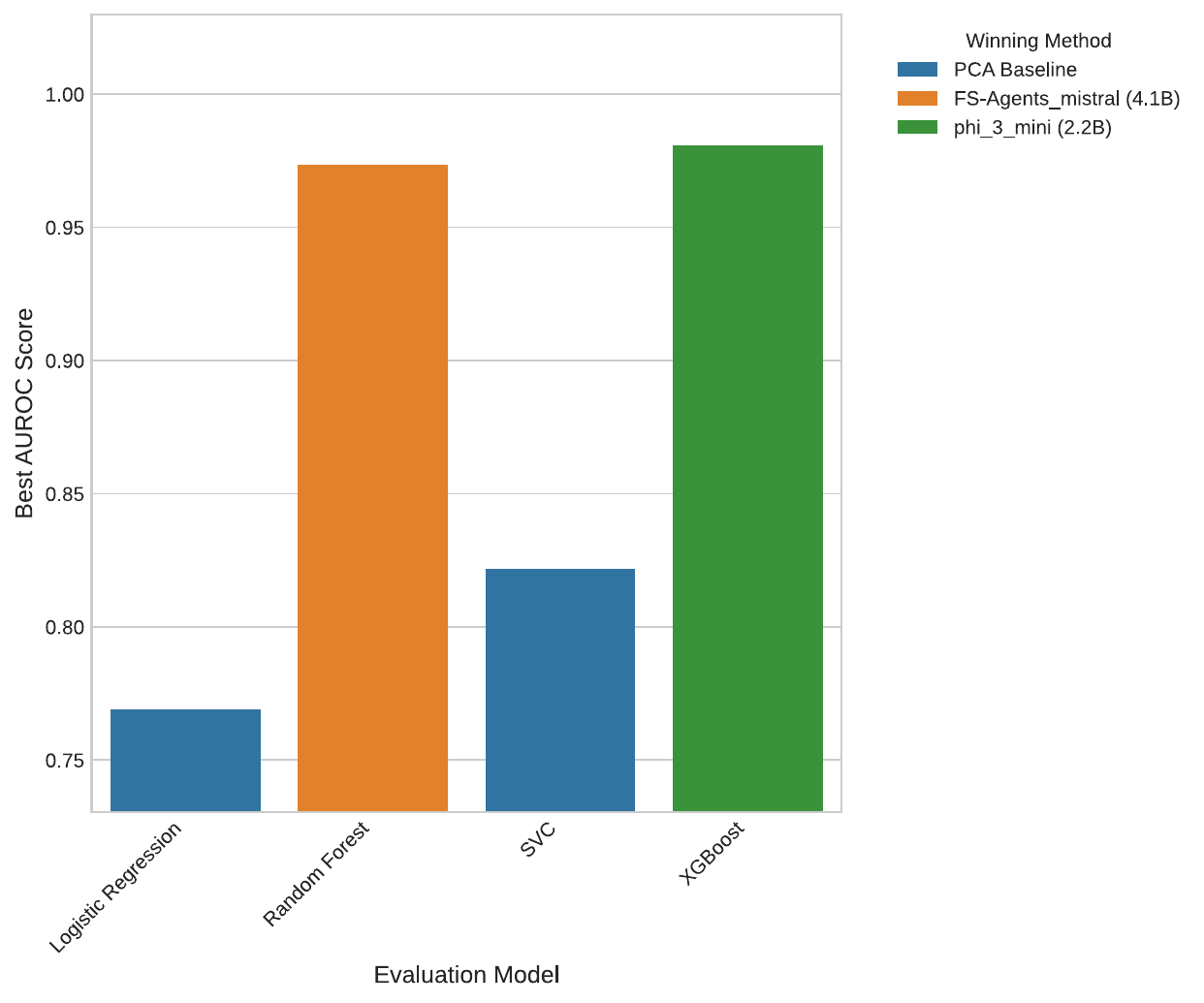}
                \caption{$n=40$}
                \label{fig:auc_n40}
            \end{subfigure}
            \hfill
            \begin{subfigure}[b]{0.32\textwidth}
                \centering
                \includegraphics[width=\linewidth]{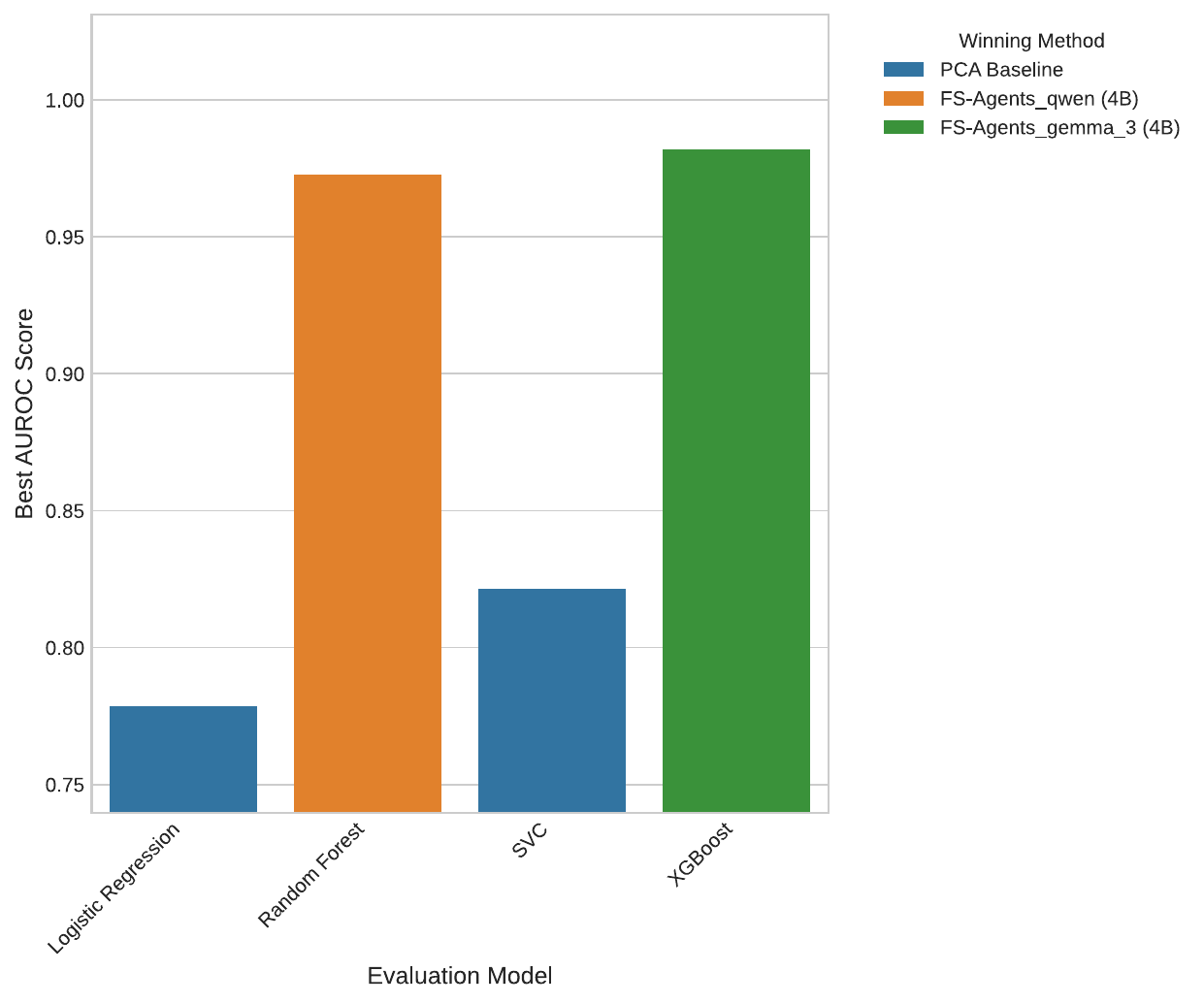}
                \caption{$n=50$}
                \label{fig:auc_n50}
            \end{subfigure}
        \end{minipage}
        \begin{minipage}[c]{0.04\textwidth} 
            \centering
            \rotatebox{90}{\textbf{AUC}}
        \end{minipage}
    \end{minipage}

    \caption{Accuracy and AUC of the best method by model for different values of ($n$)}
    \label{fig:best_method_all_metrics}
\end{figure}

\subsubsection{Performance comparison of different LLMs on feature selection}

To evaluate the impact of different LLMs on feature selection, we computed the mean AUC and accuracy for a fixed feature subset size of $n=20$, identified as optimal for most models. Table~\ref{tab:perf-auroc-acc} summarizes these results by underlying LLM architecture.
Although the overall performance difference between LLM-Select and LLM-FS-Agent is small, LLM-FS-Agent demonstrates a crucial regularizing effect on feature selection quality. When LLM-Select was initially sub-optimal (e.g. Qwen and Gemma), the deliberative approach yielded performance gains of up to $+1.49\%$ in accuracy, demonstrating its ability to correct single-agent inconsistencies and identify superior feature candidates. Conversely, slight decreases were observed for models already near their performance ceiling under LLM-Select (e.g. Mistral and Phi3-mini). These findings underscore the role of deliberation in mitigating variability inherent in single LLM outputs, enabling LLM-FS-Agent to generate more reliable feature rankings and consistently high-quality subsets.

For clarity, only the best-performing models (XGBoost and Random Forest) are reported, as SVC and LR exhibited lower and less representative performance.


\begin{table}[htbp]
\centering
\caption{Performance comparison of best-performing models on both LLM-Select and LLM-FS-Agent (metric average across different subsets)}
\label{tab:perf-auroc-acc}
\resizebox{\textwidth}{!}{%
\begin{tabular}{@{}lccccc|ccccc@{}}
\toprule
\multirow{3}{*}{\textbf{LLM Backbone}} 
& \multicolumn{5}{c|}{\textbf{AUC}} 
& \multicolumn{5}{c}{\textbf{Accuracy}} \\ 
\cmidrule(lr){2-6} \cmidrule(lr){7-11}
& \multicolumn{2}{c}{\textbf{LLM-Select}} 
& \multicolumn{2}{c}{\textbf{LLM-FS-Agent}} 
& \multirow{2}{*}{\textbf{$\Delta$\%}} 
& \multicolumn{2}{c}{\textbf{LLM-Select}} 
& \multicolumn{2}{c}{\textbf{LLM-FS-Agent}} 
& \multirow{2}{*}{\textbf{$\Delta$\%}} \\
\cmidrule(lr){2-3} \cmidrule(lr){4-5} \cmidrule(lr){7-8} \cmidrule(lr){9-10}
& \textbf{XGB} & \textbf{RF} 
& \textbf{XGB} & \textbf{RF} 
& & \textbf{XGB} & \textbf{RF} 
& \textbf{XGB} & \textbf{RF} & \\
\midrule
Mistral   & 0.9756 & 0.9639 & 0.9646 & 0.9471 & -0.99\% & 0.8920 & 0.8763 & 0.8695 & 0.8537 & -2.25\% \\
Qwen      & 0.9752 & 0.9631 & \textbf{0.9797} & \textbf{0.9741} & +0.46\% & 0.8896 & 0.8719 & \textbf{0.9017} & \textbf{0.8965} & +1.21\% \\
Gemma3-4B & 0.9733 & 0.9684 & \textbf{0.9782} & \textbf{0.9732} & +0.50\% & 0.8876 & 0.8844 & \textbf{0.9025} & \textbf{0.8936} & +1.49\% \\
LLaMA3-2B & 0.9725 & 0.9616 & \textbf{0.9740} & \textbf{0.9640} & +0.15\% & 0.8803 & 0.8638 & \textbf{0.8848} & \textbf{0.8787} & +0.45\% \\
Phi3-mini & 0.9788 & 0.9690 & 0.9724 & 0.9618 & -0.64\% & 0.9017 & 0.8856 & 0.8820 & 0.8622 & -1.97\% \\
\midrule
\textbf{Average} & 0.9751 & 0.9652 & 0.9738 & 0.9640 & +0.10\% & 0.8902 & 0.8764 & 0.8881 & 0.8769 & +0.19\% \\
\bottomrule
\end{tabular}%
}
\end{table}

\subsubsection{Computational efficiency and statistical significance}
Computational efficiency was assessed by measuring the training and inference times of the downstream classifier using the selected feature subsets. Based on the preceding results, the analysis focused on the XGBoost classifier, which demonstrated consistently high performance and, as an ensemble-based model, provides a robust yet computationally demanding baseline for evaluating the impact of feature set quality on processing speed.
Additionally, Table \ref{tab:efficiency_comparison} shows that LLM-FS-Agent substantially reduces downstream classifier training time, primarily by generating more compact feature rankings, which in turn decrease the orchestration overhead during training. Statistical analysis, reported in Table~\ref{tab:statistical_significance}, further confirms that this reduction in training time is statistically significant ($p = 0.028$) and corresponds to a large effect size (Cohen’s $d = 0.87$).

\begin{table}[htbp]
\centering
\caption{Computational efficiency comparison (XGBoost)}
\label{tab:efficiency_comparison}
\resizebox{\textwidth}{!}{%
\begin{tabular}{llccccc}
\toprule
\multirow{2}{*}{\textbf{LLM Backbone}} & \multicolumn{2}{c}{\textbf{Training Time (s)}} & \multicolumn{2}{c}{\textbf{Inference Time (s)}} & \multicolumn{2}{c}{\textbf{Speedup}} \\
\cmidrule(lr){2-3} \cmidrule(lr){4-5} \cmidrule(lr){6-7}
& \textbf{LLM-Select} & \textbf{LLM-FS-Agent} & \textbf{LLM-Select} & \textbf{LLM-FS-Agent} & \textbf{Training} & \textbf{Inference} \\
\midrule
Mistral & 0.434 & \textbf{0.108} & 0.0023 & 0.0016 & 4.02$\times$ & 1.44$\times$ \\
Qwen & 0.117 & 0.108 & 0.0014 & 0.0015 & 1.08$\times$ & 0.93$\times$ \\
Gemma3-4B & 0.101 & 0.113 & 0.0015 & 0.0016 & 0.89$\times$ & 0.94$\times$ \\
LLaMA3-2B & 0.252 & \textbf{0.108} & 0.0018 & 0.0015 & 2.33$\times$ & 1.20$\times$ \\
Phi3-mini & 0.121 & \textbf{0.117} & 0.0016 & 0.0017 & 1.03$\times$ & 0.94$\times$ \\
\midrule
\textbf{Average} & 0.205 & \textbf{0.111} & 0.0017 & 0.0016 & 1.87$\times$ & 1.09$\times$ \\
\bottomrule
\end{tabular}
}
\end{table}

\begin{table}[htbp]
\centering
\caption{Statistical significance analysis (XGBoost)}
\label{tab:statistical_significance}
\begin{tabular}{@{}lcccc@{}}
\toprule
\textbf{Metric} & \textbf{Mean Difference} & \textbf{p-value (Student’s t-test)} & \textbf{Cohen's d} & \textbf{Effect Size} \\
\midrule
AUC & +0.0010 & 0.342 & 0.28 & Small \\
Accuracy  & +0.0017 & 0.456 & 0.19 & Small \\
Training Time & -0.094s & \textbf{0.028*} & -0.87 & \textbf{Large} \\
Inference Time & -0.0001s & 0.234 & -0.31 & Small \\
\midrule
\multicolumn{5}{l}{\footnotesize *Statistically significant at $\alpha = 0.05$} \\
\multicolumn{5}{l}{\footnotesize Effect size interpretation: $|d| < 0.2$ (negligible), 0.2-0.5 (small), 0.5-0.8 (medium), $> 0.8$ (large)} \\
\bottomrule
\end{tabular}
\end{table}

\subsection{Qualitative results: final FS score and reasoning output}\label{subsec:qual_results}

The LLM-FS-Agent architecture preserves a record of the feature selection decision-making process, enabling qualitative analysis of the LLMs’ reasoning. Table~\ref{tab:feature_analysis_deliberation} provides an example of the deliberative output for two features and compares the best-performing LLM architectures, presenting their final importance scores as synthesized by the Judge, alongside the corresponding explanations.

\begin{table}[htbp]
\centering
\caption{Example of LLM-FS-Agent's deliberative analysis of Feature Importance (Final Scores and Reasoning across Qwen, Gemma, Phi3-mini, and Mistral)}
\label{tab:feature_analysis_deliberation}
\resizebox{\textwidth}{!}{%
\begin{tabular}{@{}lp{0.23\textwidth}p{0.23\textwidth}p{0.23\textwidth}p{0.23\textwidth}@{}}
\toprule
\textbf{Feature} 
& \textbf{Qwen} 
& \textbf{Gemma} 
& \textbf{Phi3-mini} 
& \textbf{Mistral} \\ 
\midrule

Src Port 
& 1.0 / The final reasoning of the arbiter is that the analysis A has a higher weight than the analysis B, and therefore, the final verdict is in favor of the analysis A. 
& 0.0 / The arbiter acknowledges the validity of both analyses but ultimately leans towards the quantitative evidence. Given the strong statistical indicators of unreliability, the final score is 0.0, recognizing source port as a weak signal in this context. 
& 0.65 / The final score leans towards Analysis A due to their comprehensive reasoning, which acknowledges both strengths and weaknesses. While source ports show limited correlation with malicious activity alone, they could provide valuable context when combined with other features as suggested by the Quantitative Refiner. 
& 0.1 / The final score is slightly biased towards Analysis B due to its recognition of specific port associations with malicious traffic, although further validation with other features is necessary. \\

\midrule

Dst Port 
& 1.0 / After carefully considering both analyses A and B, I have determined that the analysis B has a higher probability of being correct than the analysis A. Therefore, I have determined that the final score of 1.0 is based on a critical reasoning approach that takes into account counter-arguments. Therefore, the challenged score of 1.0 confirms the initial score. 
& 0.0 / The arbiter finds that while Analysis A correctly identifies the weak correlation of the destination port, Analysis B’s broader contextual perspective is more crucial for effective threat detection. Therefore, the final score is 0.0, recognizing the need for a more comprehensive analysis. 
& 0.45 / Given the moderate positive mean score of Dst Port from Quantitative Refiner's analysis, which is backed by numerical evidence despite a negative individual indicator coefficient, I find this argument more compelling. However, recognizing that no additional context was provided in Critical Challenger’s reasoning and understanding its inherent value to the discussion, it still holds significant weight. 
& 0.0 / While Dst Port can offer some indication of network traffic behavior, its weak statistical correlation with malicious activity suggests that it should not be relied upon as a definitive indicator for distinguishing between benign and malicious traffic. \\

\midrule
\multicolumn{5}{l}{\footnotesize Analyses A and B are provided by the Refiner and the Challenger, respectively.} \\
\bottomrule
\end{tabular}%
}
\end{table}

\section{Discussion}\label{sec:Discussion}

The evaluation of LLM-FS-Agent, in comparison with LLM-Select and PCA baselines, supports the hypothesis that a deliberative multi-agent framework provides a more robust, interpretable, and effective approach to feature selection. Across different classifiers and feature subset sizes, the results consistently demonstrate improvements in both predictive performance and interpretability.

The performance curves in Figure \ref{fig:performance_curves} confirm the generalizability of this approach. Across all four classifiers, the models consistently achieved high accuracy, demonstrating the effectiveness of LLM-based methods in identifying salient features for IoT intrusion detection. This result aligns with prior work on text-based feature selection, which has shown the viability of LLMs for competitive feature set prototyping. Beyond quantitative metrics, the primary contribution of LLM-FS-Agent is its introduction of transparency and structured reasoning into the feature selection process, addressing the “black box” limitations of both traditional and single-agent LLM-based methods.

The comparative analysis across various feature subset sizes ($n$) shown in Figure \ref{fig:best_method_all_metrics} reveals several key insights:

\begin{enumerate}
    \item LLM-FS-Agent's advantage in intermediate and larger subset sizes: While LLM-Select is competitive at the smallest scale ($n=5$) and outperforms at $n=40$, LLM-FS-Agent demonstrates a clear and sustained advantage at crucial intermediate and larger subset sizes ($n=10, 20, 30,$ and $50$).
    
    \item Robustness through Peer-Review: The Challenger Agent's role is particularly crucial. By actively seeking potential weaknesses, false positives, or manipulative attacker behavior, the Challenger provides an intrinsic peer-review mechanism. This process forces the final score to account for adversarial contexts, leading to a more robust selection of features.
    
    \item Robustness across LLMs and classifiers: Figure \ref{fig:best_method_all_metrics} illustrates that the best-performing architecture-LLM pair is highly conditional on the downstream classifier and the target metric. LLM-FS-Agent, however, consistently provides the most competitive performance across a wider variety of these scenarios, often leading the charts when performance is measured by both accuracy and AUC across different feature counts.
    The multi-agent approach appears to mitigate the inconsistency inherent in single-agent prompting, producing more reliable feature rankings that are less sensitive to the choice of the underlying language model or the size of the feature set.
    
    \item Optimal feature cardinality ($n=20$): The most significant performance distinction occurs at $n=20$, where LLM-FS-Agent is the only method to consistently achieve the highest results across multiple classifier and metric combinations (Figure \ref{fig:acc_n20} and \ref{fig:auc_n20}). This suggests the effectiveness of the the structured deliberation process at identifying a near-optimal subset of features that balances relevance and redundancy.

    \item Qualitative insight into feature reliability: For features like Src Port and Dst Port (Table \ref{tab:feature_analysis_deliberation}), the Judge's final reasoning reflects a nuanced, security-aware assessment justified by the understanding that these fields are easily spoofed or correlated with benign traffic. This is a critical domain-specific insight that a simple statistical ranking (filter method) or a single-agent LLM score often fails to capture, resulting in a more justified feature selection.

\end{enumerate}


The statistical analysis (Table \ref{tab:statistical_significance}) confirms LLM-FS-Agent significant advantage in reducing downstream classifier training time. This is important because it compensates for the additional overhead of the multi-agent deliberation process by selecting feature subsets that make subsequent model training more efficient. The reduction in training time is statistically significant ($p = 0.028$) and corresponds to a large effect size (Cohen’s $d = 0.87$). Although the improvements in AUC and Accuracy are not statistically significant on average, they show a small positive effect, and together with the clear efficiency gains, they support the effectiveness of the deliberative architecture.

While this work confirms the efficacy of a multi-agent deliberative process for feature selection, it is important to acknowledge certain limitations. Our experiments employed a single LLM architecture across all agent roles, specifically Llama3.2. Although this setup demonstrated clear benefits, relying on the same LLM may introduce model-specific biases. Employing different LLMs for different roles could provide greater robustness against such biases. Future research can extend this work in several promising directions. One avenue is the integration of tool-use capabilities into the agents, enabling the Refiner or Challenger to perform simple statistical tests to support their arguments with quantitative evidence. Another potential direction is the dynamic adjustment of the agent weights ($w_r, w_c$) according to the complexity or ambiguity of the feature under consideration, which may result in a more adaptive and refined deliberation process.

\section{Conclusion}\label{sec:Conclusion}

In this study, we introduce LLM-FS-Agent, a novel deliberative feature selection (FS) framework that leverages LLMs to address key limitations of interpretability and accountability in both traditional and single-agent LLM-based FS methods.
By orchestrating a structured, deliberative debate among role-specialized LLM agents (Initiator, Refiner, Challenger, and Judge), LLM-FS-Agent produces human-interpretable rationales for its feature ranking, thereby transforming feature selection into a transparent and justifiable decision-making process.
Our experiment on the CIC-DIAD 2024 intrusion detection dataset demonstrated the superiority of the proposed approach compared to the single-agent method. Specifically, the deliberative method proved highly effective in identifying a near-optimal feature subset at ($n = 20$), consistently achieving top performance across diverse downstream classifiers and evaluation metrics. These results underscore the robustness and generalizability of the feature sets selected by LLM-FS-Agent.

\section*{Declarations}
\subsection*{Acknowledgements}
The authors gratefully acknowledge Lab’Innov at Audensiel Conseil R\&D Department for providing the research environment and technical support that enabled this work. This study forms part of the advancement of the internal research and development project STAAC.


\bibliographystyle{unsrt} 
\bibliography{main}
\end{document}